\documentclass[11pt]{article}
\PassOptionsToPackage{authoryear}{natbib}
\usepackage[final]{neurips_2024}
\usepackage[none]{hyphenat}
\usepackage{hyperref} 
\usepackage{authblk} 
\usepackage{booktabs}
\usepackage{adjustbox}
\usepackage{graphicx}
\usepackage{amsfonts}
\usepackage{multicol}
\usepackage{multirow}
\usepackage{tabularx}
\usepackage{enumitem,amsmath,bm,amssymb}
\usepackage{pifont,makecell}
\usepackage{tikz}
\usepackage{CJKutf8}
\usepackage{colortbl}
\usepackage{array}
\usepackage{comment}
\usepackage{geometry}
\usepackage{verbatim}
\usepackage{tcolorbox}
\usepackage{fontawesome}
\usepackage{booktabs}
\usepackage{subcaption}

\title{MinMo: A Multimodal Large Language Model for Seamless Voice Interaction}

\author{FunAudioLLM Team}

\affil{Tongyi Lab, Alibaba Group}
\affil{FunAudioLLM@list.alibaba-inc.com}
\date{} 
\begin{document}

\maketitle

\begin{abstract}
Recent advancements of large language models (LLMs) and subsequent multimodal speech-text models have provided promising foundation technologies for achieving seamless voice interactions, that is, real-time, natural, smooth, and human-like voice conversations between the user and the system.  Prior works of speech-text multimodal models for voice interactions can be roughly categorized into \textit{native} and \textit{aligned} models. Native multimodal models simultaneously model end-to-end understanding and generation of both speech and text with a single framework; however, they face the challenges of drastic discrepancy between speech and text sequence lengths, insufficient speech pre-training, and catastrophic forgetting of knowledge of text LLMs. Aligned multimodal models are more successful at maintaining capabilities of text LLMs; yet existing models are usually trained on small-scale speech data, investigated on a limited set of speech tasks, and lack systematic exploration of instruction-following capabilities for rich and nuanced speaking styles. In this work, we introduce \textbf{MinMo}, a Multimodal Large Language Model with approximately 8B parameters for seamless voice interaction. We address the main limitations of prior \textit{aligned} multimodal models. We train MinMo through multiple stages of speech-to-text alignment, text-to-speech alignment, speech-to-speech alignment, and duplex interaction alignment, on 1.4 million hours of diverse speech data and a broad range of speech tasks. After the multi-stage training, MinMo achieves state-of-the-art performance across various benchmarks for voice comprehension and generation while maintaining the capabilities of text LLMs, and also facilitates full-duplex conversation, that is, simultaneous two-way communication between the user and the system. Moreover, we propose a novel and simple voice decoder that outperforms prior models in voice generation. The enhanced instruction-following capabilities of MinMo supports controlling speech generation based on user instructions, with various nuances including emotions, dialects, and speaking rates, and mimicking specific voices. For MinMo, the speech-to-text latency is approximately 100ms, full-duplex latency is approximately 600ms in theory and 800ms in practice. The MinMo project web page is \url{https://funaudiollm.github.io/minmo}, and the code and models will be released soon. 

\end{abstract}

\section{Introduction}
\label{sec:intro}

\begin{figure*}[ht]
  \centering
  \includegraphics[width=0.8\linewidth]{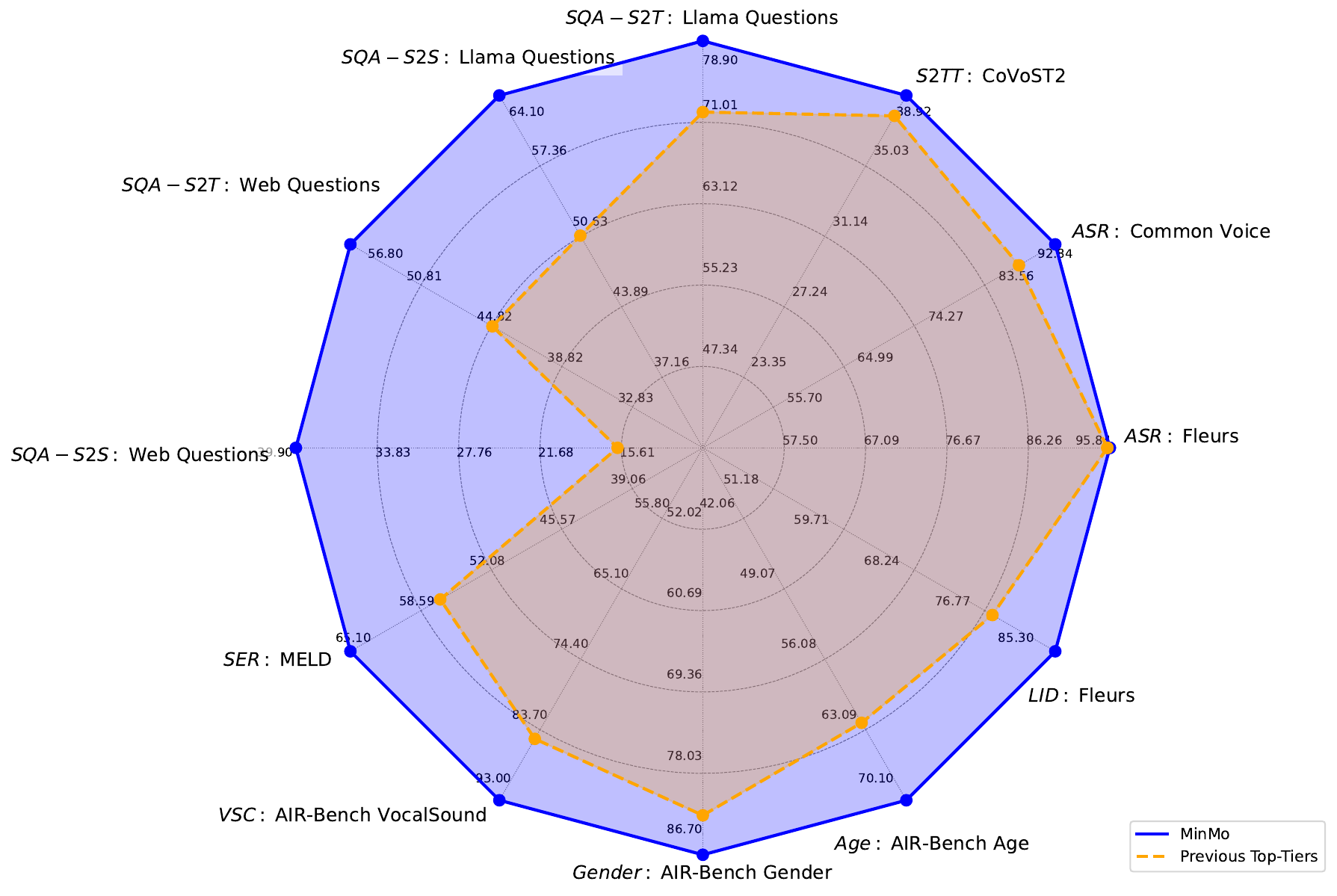}
  \caption{Performance comparison between our \textbf{MinMo}($\sim$8B parameters) and top-tier speech-text multimodal models, including Moshi(7B)~\citep{DBLP:journals/corr/abs-2410-00037}, Freeze-Omni(7.5B)~\citep{wang2024freeze}, GLM-4-Voice(9B)~\citep{zeng2024glm}, SeamlessM4T Large v2(2.3B)~\citep{DBLP:journals/corr/abs-2308-11596}, NExT-GPT(12.42B)~\citep{DBLP:conf/icml/Wu0Q0C24}, speech-to-text model Qwen2-Audio($\sim$8B)~\citep{chu2024qwen2}, Whisper-large-v3(1.55B)~\citep{radford2023robust}, and others. We demonstrate capabilities of MinMo on automatic speech recognition (ASR), speech-to-text translation (S2TT), spoken question answering (SQA) encompasses both speech-to-text (S2T) and speech-to-speech (S2S), vocal sound classification (VSC), speech emotion recognition (SER), language identification (LID), age recognition and gender detection. ASR is evaluated using 1-WER\%, with Fleurs \& Common Voice results are averaged over 10 languages (\textit{zh, en, ja, ko, yue, de, fr, ru, es, it}). S2TT is evaluated using BLEU, with CoVoST2 results averaged over \textit{en2zh, en2ja, {zh/ja/de/fr/ru/es/it}2en} translation directions. SQA is eavaluated using Accuracy. SER is evaluated using Weighted Accuracy. \textbf{MinMo surpasses the previous SOTA models on all these tasks.}}
  \label{fig:benchmark_radar}
\end{figure*}

Seamless voice interaction indicates that \textit{a user experiences real-time, natural, relevant, and human-like spoken conversation with the system}.
Facilitating seamless voice interaction poses great challenges: (1) the system needs to understand audio accurately and comprehensively, including comprehending the content and also paralinguistic cues in speech (e.g., emotion, prosody) as well as audio events; (2) the system is expected to produce natural and expressive speech response; (3) the system should provide relevant and reasonable response to the user, as an intelligent chatbot; (4) the system is expected to support full-duplex conversation (simultaneous two-way communication), that is, the system listens while speaking and the user is free to interrupt when the system is speaking, then the system either continues the speech, or concedes it, listens to the user, and provides response to the new user query.

In recent years, seamless voice interaction systems have gained significant momentum, especially with the advancements in multimodal large language models, such as GPT-4o~\citep{hurst2024gpt} and Moshi~\citep{DBLP:journals/corr/abs-2410-00037}. These systems not only produce natural and expressive speech but also understand cues beyond words, including emotional tones and audio events. 
Current multimodal language models for voice interaction can be categorized into two main categories. 
The first category includes \textit{native multimodal models}, such as Moshi~\citep{DBLP:journals/corr/abs-2410-00037} and GLM-4-Voice~\citep{zeng2024glm}.  These models typically use a decoder-only Transformer as the backbone to simultaneously model understanding and generation of both speech and text modalities within a single framework; they usually require pre-training with both speech and text data. These models suffer from two major limitations.  Firstly, after speech discretization, speech token sequences are often more than twice the length of text (e.g., 12.5 tokens per second in Moshi). This discrepancy in sequence length poses challenges as model sizes grow, such as the 175B GPT-3~\citep{DBLP:conf/nips/BrownMRSKDNSSAA20}. Secondly, the scarcity of speech data compared to text leads to highly imbalanced speech-text training data and in turn causes catastrophic forgetting~\citep{wang2024freeze}.

The second category includes \textit{aligned multimodal models}, integrating voice capabilities while aiming to maintain the capabilities of the existing pre-trained text LLM. This results in intermediate outputs that still contain text, as seen in models such as Llama-Omni~\citep{DBLP:journals/corr/abs-2409-06666} and Freeze-Omni~\citep{wang2024freeze}. However, these alignment-based models are typically trained on limited speech data (200K samples for LLaMA-Omni and 120K hours for Freeze-Omni), leading to questions on the impact of larger speech datasets on model capabilities and whether the chat capabilities of the original text-LLM might be compromised. Furthermore, investigation of extensive speech tasks has not been conducted on these models, such as speech translation, emotion recognition, speaker analysis, language identification, and audio event detection. Moreover, these models lack systematic evaluations of instruction-following capabilities for rich and nuanced speaking styles, as well as lacking development and evaluation of full-duplex conversation capabilities, for achieving seamless voice interaction.

In this work, we introduce a new multimodal large language model \textbf{MinMo}, to address these limitations of existing aligned multimodal models. MinMo is trained on over 1.4 million hours of speech data, encompassing various tasks such as Speech-to-Text, Text-to-Speech, and Speech-to-Speech, as detailed in Table~\ref{tab:MinMo_data}. This extensive training enables MinMo to achieve state-of-the-art (SOTA) performance across various benchmarks, as shown in Figure~\ref{fig:benchmark_radar}. We also apply methods that effectively mitigate catastrophic forgetting of the chat capabilities of the original text-LLM while enhancing voice comprehension and generation after training on such large-scale datasets.

We also propose a novel voice decoder that balances structural simplicity and competitive voice generation performance. LLaMA-Omni uses a non-autoregressive (NAR) streaming Transformer, which takes the output hidden states of the LLM as input and employs connectionist temporal classification (CTC) to predict the discrete speech token sequence of the response. This approach suffers from inferior performance compared to autoregressive speech decoder. Freeze-Omni uses three speech decoders, including NAR prefix speech decoder, NAR speech decoder, and AR speech decoder, which complicates the model structure. Different from both of these strategies, we design an AR streaming Transformer for MinMo, which mixes the output hidden states of the LLM with speech tokens, based on a fixed ratio, as shown in Figure~\ref{fig:MinMo}.

Our contributions can be summarized as follows:

\begin{itemize}[leftmargin=*,noitemsep]
    \item We propose \textbf{MinMo}, an end-to-end aligned multimodal large model that gains audio understanding, audio generation, and end-to-end duplex speech interaction capabilities by adapting a pre-trained text large language model (LLM) through a multi-stage alignment strategy over 1.4 million hours of audio data covering a wide range of speech tasks. MinMo achieves state-of-the-art (SOTA) performance on multiple open-source benchmarks, including spoken dialogue, multilingual speech recognition, speech translation, emotion recognition, and speaker analysis. Different from previous multimodal models that often suffer from notable catastrophic forgetting of capabilities of the text LLM and significant performance degradation on text tasks, MinMo has minimal loss in the original capabilities of the text LLM. 
    
    \item We propose a novel alignment method for streaming end-to-end audio generation, by exploring the use of the hidden layer representations of the text model as inputs to the Voice Decoder for aligning the audio output modality. Experimental results demonstrate that our streaming voice decoder effectively balances structural simplicity, low latency, and high voice generation performance, and outperforms previous models. Additionally, while most existing voice interaction systems only support controlling the content of the response, MinMo enhances instruction-following capabilities and enables the generation of speech corresponding to user-specified emotions, dialects, and speaking rates, as well as mimicking specific voices with a 98.4\% instruction-following accuracy.

    \item We develop a mechanism that effectively facilitates full-duplex interactions with MinMo. Specifically, we implement a full-duplex prediction module that harnesses the text LLM's semantic understanding capabilities to decide whether to continue system response, or concede, listen, and respond to new user query. For MinMo, the speech-to-text latency is approximately 100ms; the full-duplex latency is approximately 600ms in theory and 800ms in practice. 
\end{itemize}

\begin{figure*}[ht]
    \centering
    \begin{subfigure}{0.45\textwidth}
        \centering
        \includegraphics[width=\textwidth]{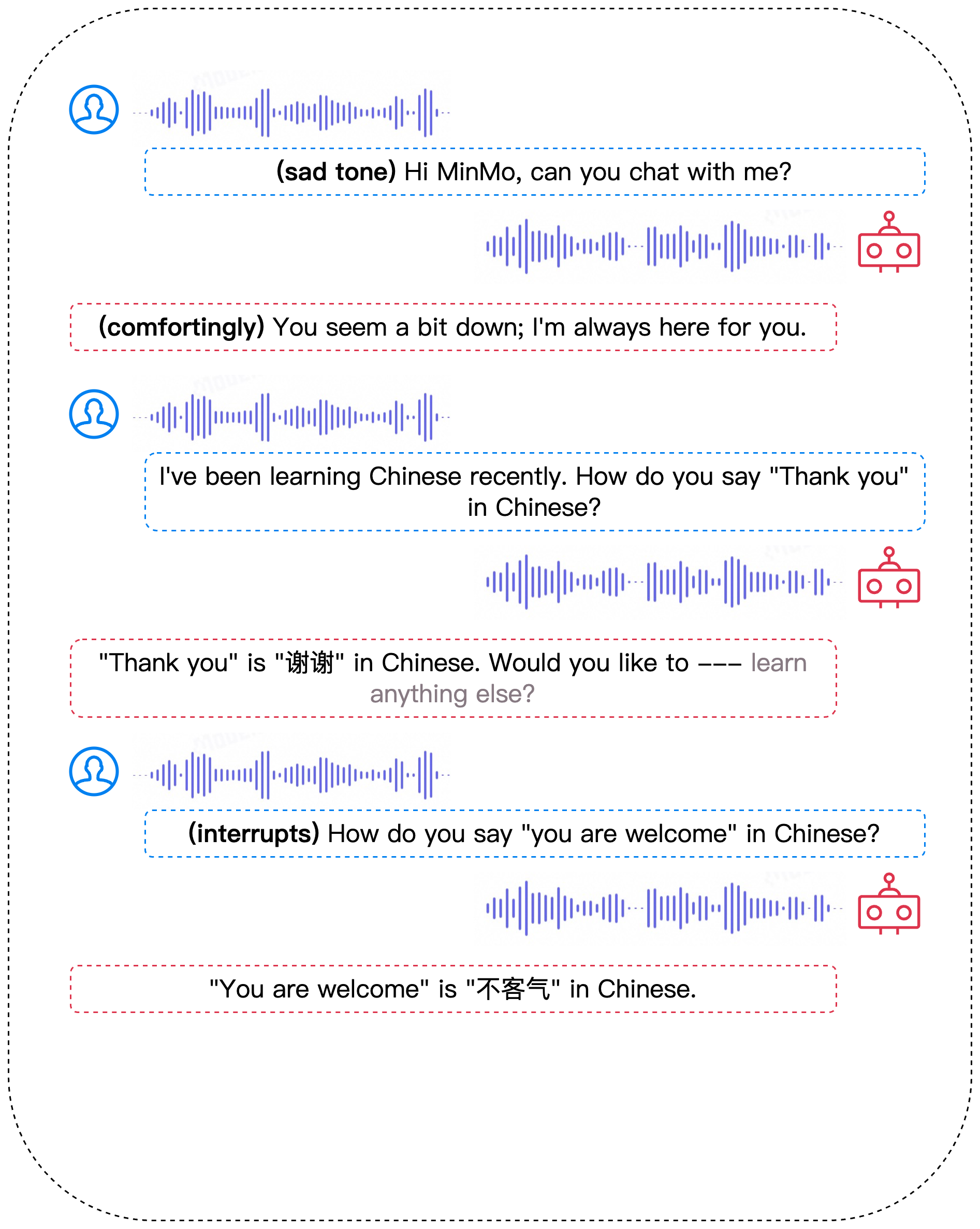}
        \caption{An example showcases MinMo's capabilities, including speech-to-speech chat, speech-to-text translation, style-controllable speech synthesis, and full duplex interaction.}
        \label{fig:sub1}
    \end{subfigure}
    \hfill
    \begin{subfigure}{0.45\textwidth}
        \centering
        \includegraphics[width=\textwidth]{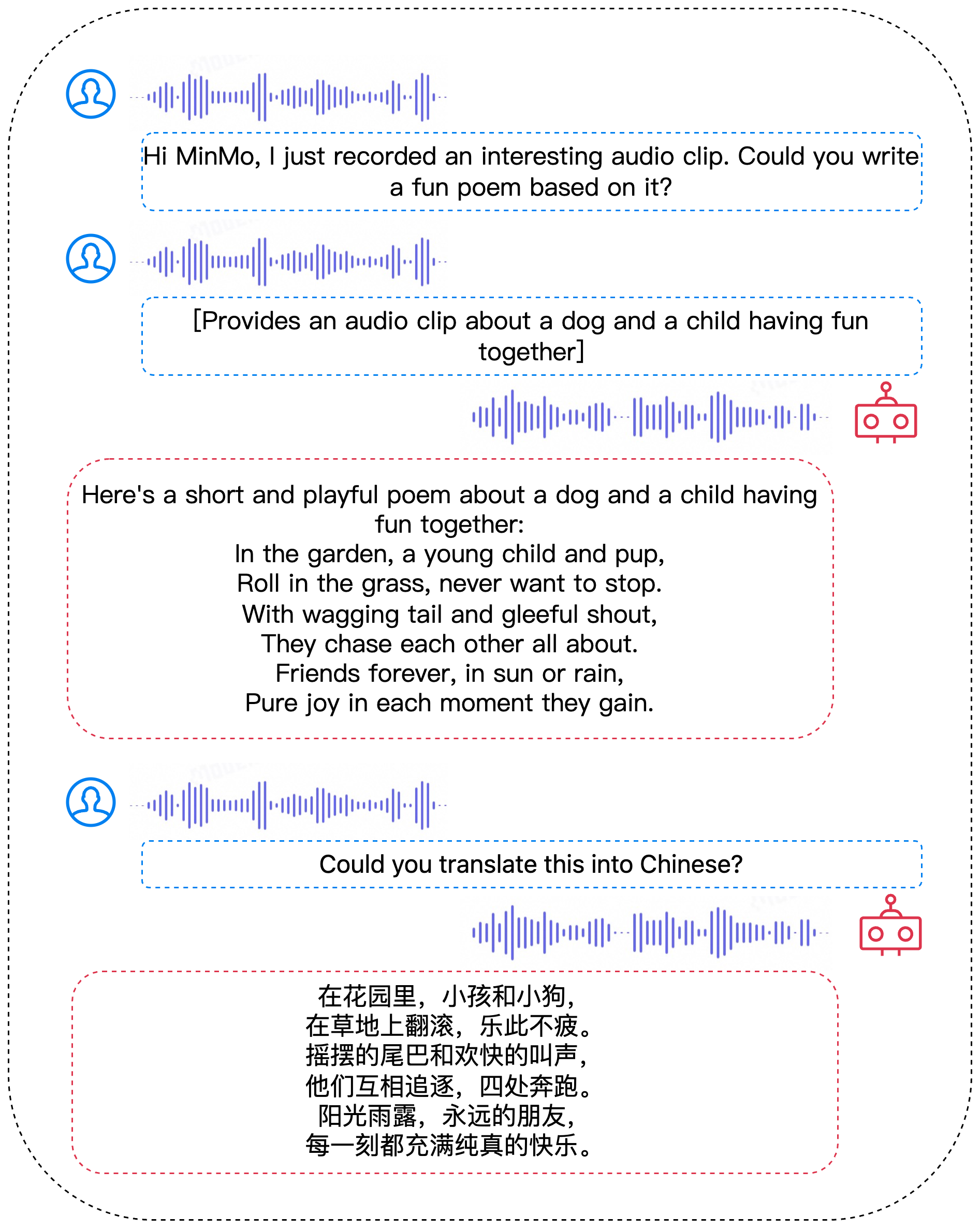}
        \caption{An example showcases MinMo's capabilities, including speech-to-speech chat, audio event detection, speaker analysis and speech-to-text translation.}
        \label{fig:sub2}
    \end{subfigure}
    \caption{Examples demonstrating various capabilities of MinMo. More capabilities of MinMo include the tasks shown in Table~\ref{tab:MinMo_data}.}
    \label{fig:main}
\end{figure*}

\section{Related Work}

\paragraph{Multimodal Spoken Dialogue Models}
A variety of speech foundation models have been developed for generic audio understanding, but not systematically explored for voice interaction. For example, Qwen2-Audio \citep{chu2024qwen2} integrates Whisper speech encoder with a pre-trained text LLM and adapts the LLM for speech understanding capabilities through multi-task pre-training and instruction-based supervised fine-tuning. SALMONN~\citep{salmonn} is another speech-text LLM for generic audio understanding, by integrating separate speech and audio encoders with a pre-trained text LLM through Q-Former and adopting LoRA for modality alignment.

Since this work aims to develop an end-to-end multimodal model for seamless voice interaction, we focus on comparing MinMo to speech-text models for voice interaction (or called multimodal spoken dialogue models). Contemporaneously or inspired by GPT-4o, there have been active developments of multimodal spoken dialogue models managing to achieve real-time voice conversations with user. \citep{DBLP:journals/corr/abs-2411-13577} provides an in-depth overview of recent spoken dialogue models. Some works support traditional turn-based voice chat (i.e., half-duplex communication), but cannot handle full-duplex voice interaction (i.e., simultaneous two-way communication). These models include collaborative systems and end-to-end frameworks. PSLM~\citep{DBLP:conf/emnlp/MitsuiMWHS24} is a collaborative system since it replies on ASR to process audio input, which discards paralinguistic information and causes error propagation. PSLM generates speech and text tokens in parallel hence it reduces the speech generation latency; however, it suffers from reduced response quality.  Different from the collaborative systems such as PSLM, end-to-end frameworks directly accept audio input and generate audio output.  Llama-Omni~\citep{DBLP:journals/corr/abs-2409-06666} and Mini-Omni~\citep{xie2024mini} are two recent end-to-end frameworks that have not been trained for full-duplex communication. Llama-Omni integrates Whisper speech encoder, speech adapter, streaming speech decoder, and vocoder with a pre-trained text LLM backbone. The speech decoder generates discrete units corresponding to generated text prefix in an NAR manner. The model is trained with a two-stage strategy: in the first stage, the speech encoder is frozen, and the speech adapter and LLM are trained autoregressively; in the second stage, the speech encoder, speech adapter, and LLM are frozen and only the speech decoder is trained using the CTC loss. Llama-Omni is evaluated on speech-to-text instruction-following and speech-to-speech instruction-following tasks.  Mini-Omni also adopts Whisper encoder and uses adapter for minimal training in order to reserve LLM's capabilities. The model is trained through three stages of modality alignment, adapter training, and multi-modal fine-tuning. Mini-Omni simultaneously generates text and audio tokens, while padding N tokens to ensure that the corresponding text tokens are produced first to guide audio token generation. 

Our MinMo facilitates full-duplex spoken dialogues. Existing full-duplex voice chat systems can also be categorized into collaborative systems and end-to-end models. Among collaborative systems, VITA~\citep{DBLP:journals/corr/abs-2408-05211} runs two models at the same time, namely, the generation model and the monitoring model, to support full-duplex communication. When the generation model is generating system response, the monitoring model monitors the environment and once it detects effective user interruption, it combines context and provides response to the new user query, while the generation model pauses and switches to the monitoring role. Notably, VITA still relies on an external TTS module to generate speech output.  Alternatively, another collaborative system~\citep{DBLP:journals/corr/abs-2405-19487} operates with LLM interfacing with an ASR module and a streaming TTS module. The system does not require modality alignment; instead,  supervised fine-tuning is conducted on a pre-trained text LLM with the following paradigm: At each time step, the LLM either processes an input token, or generates a text token, or outputs a special control token for state transitions between SPEAK and LISTEN. All these tasks are defined as next token prediction on a serialized, single-stream view of dialogues. Full-duplex dialogue learning is conducted on data synthesized by GPT-4 to generate dialogues with different types of user interruptions. Notably, with its cascaded architecture, this system suffers from high latency up to 680ms. 

Among end-to-end full-duplex models, the early work of dGSLM~\citep{DBLP:journals/corr/abs-2203-16502} proposes a Siamese architecture to jointly process both audio token streams of user speech and system speech. However, it suffers from several weaknesses: it relies on speech-only training, hence does not leverage capabilities of a pre-trained text LLM; it only uses semantic tokens, hence does not sufficiently model acoustic information; it does not support online mode. LSLM~\citep{DBLP:journals/corr/abs-2408-02622} uses a decoder-only Transformer to generate speaking tokens and a streaming SSL encoder to process listening tokens. It introduces an interruption token to stop speaking when detecting a turn-taking attempt from the user. However, the model is insufficient in generating reasonable responses. Among the more recent end-to-end full-duplex models, Moshi~\citep{DBLP:journals/corr/abs-2410-00037}, GLM-4-Voice~\citep{zeng2024glm}, SyncLM~\citep{DBLP:conf/emnlp/VeluriPYGG24}, IntrinsicVoice~\citep{DBLP:journals/corr/abs-2410-08035}, and Omni-Flatten~\citep{DBLP:journals/corr/abs-2410-17799} are \textit{native multimodal models}. They simultaneously model understanding and generation of both speech and text modalities within a single framework, based on a GPT backbone, and require self-supervised autoregressive pre-training using both speech and text data. As discussed in Section~\ref{sec:intro}, these native multimodal models need to tackle the challenges due to significant discrepancy between sequence lengths of speech tokens and text tokens, and also highly imbalanced speech-text training data and the resulting catastrophic forgetting.  
IntrinsicVoice employs a GroupFormer to generate HuBERT tokens from the LLM’s hidden states, effectively shortening speech sequences to lengths comparable to text sequences.
OmniFlatten utilizes a multi-stage progressive post-training strategy that incorporates a chunk-based flattened single stream of speech tokens and text tokens to learn full-duplex and text-free speech-to-speech interaction.
Different from these native multimodal models, our MinMo is in the category of \textit{aligned multimodal models}, which also include Llama-Omni, Mini-Omni2 \citep{xie2024mini}, and Freeze-Omni~\citep{wang2024freeze}. Aligned multimodal models integrate voice capabilities while aiming to maintain the capabilities of the existing pre-trained text LLM. Mini-Omni2 introduces a command-based interruption mechanism for supporting full-duplex conversation; however, it is only evaluated on the ASR task and compared to Whisper, VITA, and Mini-Omni.  Freeze-Omni~\citep{wang2024freeze} is a speech-to-speech model that freezes the pre-trained text LLM to reserve the LLM's capabilities. It supports streaming input speech and generates streaming output speech, uses multi-task training, and conducts chunk-level state prediction for modeling full-duplex voice interaction. Our MinMo differs from these aligned multimodal models in the following ways. We explore training MinMo on much larger speech datasets (1.4 million hours of diverse speech data in contrast to 200K samples for LLaMA-Omni and 120K hours for Freeze-Omni) and on much more extensive speech tasks. MinMo also differs from existing aligned multimodal models with a novel speech decoder, enhanced instruction following capabilities, and systematic training and evaluation of full-duplex spoken conversation capabilities.

\paragraph{Text Style-Controllable Speech Synthesis}
A distinctive feature of multimodal spoken dialogue models, compared to text-based dialogue models, is their ability to comprehend and generate acoustic information beyond mere textual content. The speech modality not only contains the content but also acoustic information such as emotion, dialect, and speaking rate. An intelligent multimodal spoken dialogue model should be able to comprehensively understand the acoustic information in input speech (e.g., emotion) and also ideally generate responses with specified emotions, dialects, and speaking rate, as well as mimicking specific voices, so that the system can achieve a deeper level of understanding and response in communication. Collaborative systems, such as ParalinGPT~\citep{DBLP:conf/icassp/LinSGYGGSLB24}, E-Chat~\citep{DBLP:journals/corr/abs-2401-00475}, and Spoken-LLM~\citep{DBLP:conf/acl/LinCL24}, incorporate paralinguistic features to enhance the understanding of acoustic information such as emotions. These systems can be cascaded with a style-controllable Text-to-Speech (TTS) system to generate responses with specific emotion, speaking rate, and volume. Significant progresses have been made in text-style controllable TTS, such as TextrolSpeech~\citep{DBLP:conf/icassp/JiZ00CDHZ24}, PromptTTS~\citep{DBLP:conf/icassp/ShimizuYKSDKT24}, PromptTTS2~\citep{DBLP:conf/iclr/LengGSJ0LLYZS0024}, InstructTTS~\citep{DBLP:journals/taslp/YangLHWM24}, and ControlSpeech~\citep{DBLP:journals/corr/abs-2406-01205}. In contrast to these collaborative systems, Moshi~\citep{DBLP:journals/corr/abs-2410-00037} uses a TTS engine with a single actor's voice and recorded monologues in over 70 speaking styles to synthesize training data to support understanding and generation of acoustic information in an end-to-end model. GLM-4-Voice~\citep{zeng2024glm} employs high-quality, multi-turn spoken dialogues tailored to specific speech style requirements, such as speaking rate, emotion, or dialect, to support style-controllable spoken dialogues. However, to the best of our knowledge, no previous work has demonstrated that aligned multimodal models can support style-controllable voice generation. Contrary to previous claims that aligned multimodal models such as Llama-Omni and Freeze-Omni only allow language models to control the content of speech but not the style and prosody~\citep{zeng2024glm}, in this work, we propose a novel streaming voice decoder for the aligned multimodal model MinMo and find that this decoder enhances instruction-following capabilities and enables MinMo to generate speech corresponding to user-specified emotions, dialects, speaking rates,  as well as mimicking specific voices.

\section{MinMo}

\subsection{Model Architecture}
\begin{figure*}[t!]
  \centering
  \includegraphics[width=1.0\linewidth]{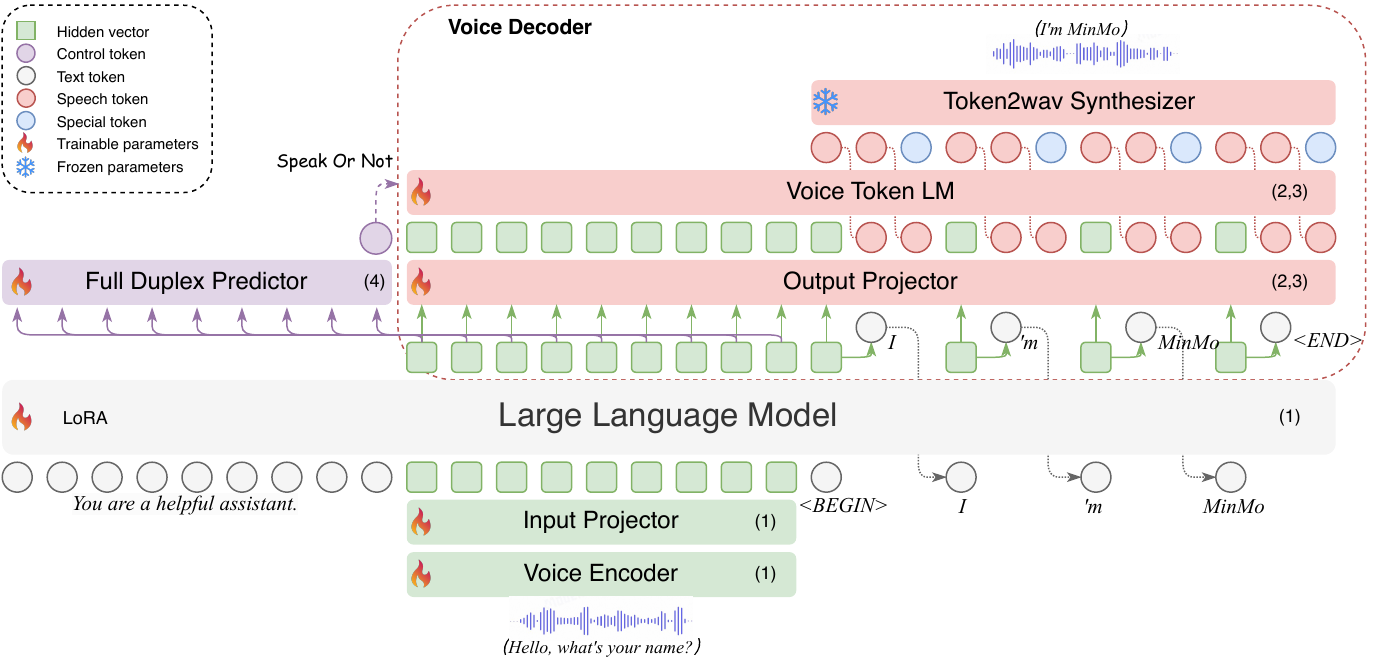}
  \caption{The overall architecture of MinMo.  Table~\ref{tab:MinMo_module} provides detailed descriptions of each module in this diagram.}
  \label{fig:MinMo}
\end{figure*}

Figure~\ref{fig:MinMo} illustrates the model architecture of MinMo.  MinMo employs a lightweight modality alignment approach on a pretrained text LLM. 
Table~\ref{tab:MinMo_module} provides detailed descriptions of each module in MinMo.

The \textit{Voice Encoder} is initialized with the pretrained SenseVoice-large encoder module \citep{DBLP:journals/corr/abs-2407-04051}, which provides robust voice understanding capabilities and supports multilingual speech recognition, emotion recognition, and audio event detection.
The \textit{Input Projector} consists of a randomly initialized two-layer Transformer combined with a CNN layer for dimensional alignment and downsampling.  We use the pretrained Qwen2.5-7B-instruct model~\citep{qwen2.5}\footnote{\url{https://huggingface.co/Qwen/Qwen2.5-7B-Instruct}} as the pre-trained text LLM, due to its outstanding performance on various benchmarks~\citep{qwen2.5}.
We utilize the streaming audio generation mechanism of CosyVoice 2~\citep{du2024cosyvoice2scalablestreaming}, due to its low latency and competitive speech synthesis performance. For every batch of five text tokens received, we pass these tokens and their corresponding final hidden layer vectors simultaneously to the \textit{Output Projector} and the \textit{Voice Token LM}. The \textit{Output Projector} is a single-layer linear module randomly initialized for dimensional alignment. The \textit{Voice Token LM} uses the pretrained CosyVoice 2 LM module. 
The \textit{Voice Token LM} then autoregressively generates fifteen speech tokens, ensuring efficient and seamless audio synthesis. These audio tokens are processed in real time by the \textit{Token2wav Synthesizer} module to produce the final audio output. 
The \textit{Token2wav Synthesizer} comprises a pretrained flow-matching model, which converts tokens to mel spectrograms, and a pretrained vocoder, which transforms mel spectrograms into waveforms, both sourced from CosyVoice 2.
MinMo is fully trained end-to-end using additional hidden embeddings, which facilitate control of speech styles, such as emotion, dialect, and speaking rate, based on user instructions. Details of voice generation are elaborated in Section~\ref{sec:voice_decoder}. The \textit{Full Duplex Predictor} module 
consists of a single-layer Transformer and a linear softmax output layer, both randomly initialized.  This module performs real-time prediction on whether to engage with user commands or temporarily halt the ongoing system broadcast to allow for processing further audio input from the user. Once the Full Duplex Predictor decides that a system response is appropriate, MinMo produces text outputs and concurrently generates the audio tokens in a token-by-token manner. 

MinMo has approximately 8 billion parameters in total. The training procedure of MinMo is detailed in Section~\ref{sec:minmo_training}.
The end-to-end latency from receiving the user's audio input to delivering the audio response is approximately 600 ms, when tested on the L20 GPU. 

\subsection{Streaming Voice Decoder}
\label{sec:voice_decoder}
To facilitate natural voice responses for MinMo, we introduce a novel voice decoder that transforms textual outputs from an LLM into speech. As illustrated at the top of Figure~\ref{fig:MinMo}, our voice decoder comprises three components: \textbf{an output projector}, \textbf{a voice token language model (LM)}, and \textbf{a streaming token-to-wave (token2wav) synthesizer}.

The output projector aligns the dimensions of the LLM with those of the voice decoder. The hidden states from the LLM contain rich contextual information but are semantically ambiguous; whereas, the sampled text tokens are more precise and consistent with the generated text.  Meanwhile, the hidden states of the current round of user input contain explicit instruction information. 
For every dialog turn, the embeddings of user input, and hidden states of the LLM’s last layer output will be concatenated along the feature dimension to form the query embeddings. The query embeddings, and embeddings of five sampled text tokens along with the hidden states of the LLM’s last layer output will be concatenated along the sequence dimension and fed into the projector. 
In this report, the projector’s outputs are referred to as \textbf{semantic vectors}, which represent rich and accurate semantic information. 

Following the output projector, a voice token LM is employed to generate speech tokens autoregressively. This LM operates on sequences interleaving text and speech tokens. Specifically, we intermix the semantic vectors and speech tokens in a fixed ratio of 5:15, that is,  every five semantic vectors are followed by fifteen speech tokens. During training, a teacher forcing strategy is applied, and a special token is introduced to signal that the next semantic vectors should be concatenated. Once the LLM's textual response is complete and the semantic vectors are exhausted, we insert a ``turn of speech'' token to signal the voice token LM that subsequent tokens should be speech tokens exclusively. The speech synthesis process concludes when the ``end of speech'' token is generated.

For reconstructing waveforms from the speech tokens, we utilize an off-the-shelf streaming token2wav synthesizer, as described by~\citet{du2024cosyvoice2scalablestreaming}. The token2wav synthesizer incorporates a chunk-aware flow matching model and a mel-to-wave vocoder, capable of synthesizing waveforms in chunks of fifteen tokens. 

The theoretical latency of the voice decoder can be computed as follows:
\begin{align}
Latency = 5d_{llm} + 15d_{lm} + 15d_{syn}
\end{align}
where $d_{llm}$ denotes the computation time for the LLM to generate one text token, $d_{lm}$ denotes the time for the LM to generate one speech token, and $d_{syn}$ denotes the time for the token2wav synthesizer to generate the waveforms corresponding to each speech token.

\begin{table}[htbp]
    \centering
\begin{tabular}{>{\raggedright\arraybackslash}p{3.5cm}>{\raggedright\arraybackslash}p{6.2cm}>{\centering\arraybackslash}p{2.7cm}}
\toprule
\textbf{Module} & \textbf{Description} & \textbf{Number of Parameters} \\ 
\midrule
Voice Encoder & Initialized with the encoder parameters of the pre-trained SenseVoice-Large audio understanding model \citep{DBLP:journals/corr/abs-2407-04051} & $\sim636$M \\ 
\midrule
Input Projector & 2 Transformer layer and 1 CNN layer for dimensional transformation and perform 2x downsampling on the input & $\sim170$M \\ 
\midrule
Large Language Model & Initialized with Qwen2.5-7B-instruct \citep{qwen2.5} & 7B \\ 
\midrule
Output Projector & Linear layer for dimensional transformation & $\sim6$M \\ 
\midrule
Voice Token LM & Initialized with the LLM of the pre-trained CosyVoice2 \citep{du2024cosyvoice2scalablestreaming} & $\sim370$M \\ 
\midrule
Full Duplex Predictor &   1 Transformer layer and 1 linear-softmax output layer, both randomly initialized  & $\sim18$M \\
\bottomrule
\end{tabular}
\vspace{3mm}
    \caption{Descriptions of the modules in MinMo as depicted in Figure~\ref{fig:MinMo}. MinMo has approximately 8 billion parameters in total.
    }
    \label{tab:MinMo_module}
\end{table}

\subsection{Tasks and Training Data}
\begin{table}[htbp]
    \centering
\begin{tabular}{llc}
\toprule
\textbf{Category} & \textbf{Specific Tasks} & \textbf{Hours} \\ 
\midrule
\multirow{9}{*}{\textbf{Speech-to-Text}}                   
                   &   Automatic Speech Recognition (ASR) & 630k\\
                   &   Speech-to-Text Translation (S2TT)  &  451k \\
                   &   Language Identification (LID)           &  34k\\
                   &   Contextual Bias Speech Recognition                  &  50k\\
                   &   Speech Emotion Recognition (SER)       & 48k\\
                   &   Audio Event Detection (AED)         & 11k\\
                   &   Speaker Analysis    & 24k\\
                   &   Spoken Language Smoothing         & 0.4k \\ 
                   &   Speech-to-Text Chat                  & 10k \\
\midrule
\multirow{2}{*}{\textbf{Text-to-Speech}}        &   Speech Synthesis               &  170k   \\
                   &   Instruct Speech Synthesis      &  1k  \\
\midrule
\multirow{2}{*}{\textbf{Speech-to-Speech}}   &   Speech-to-Speech chat      &  10k  \\
                   &  Style-controllable Speech-to-Speech Chat       &  0.1k  \\                   
\midrule
\textbf{Speech-to-ControlToken} &   Full Duplex Interaction    & 4k \\
\bottomrule
\end{tabular}
\vspace{3mm}
    \caption{The multitask training data for MinMo. Task specifications can be found in Section~\ref{sec:experiments}.}
    \label{tab:MinMo_data}
\end{table}

The training tasks for MinMo consist of four categories, including \textit{Speech-to-Text}, \textit{Text-to-Speech}, \textit{Speech-to-Speech}, and \textit{Speech-to-ControlToken} tasks. The specific tasks within each category and their corresponding data scales are presented in Table~\ref{tab:MinMo_data}.

\textbf{Speech-to-Text tasks.} This category consists of approximately 1.2 million hours of speech-text paired data, including tasks such as automatic speech recognition (ASR), speech-to-text translation (S2TT), language identification (LID), contextual biasing speech recognition, speech emotion recognition (SER), audio event detection (AED), speaker analysis, spoken language smoothing.
The training data for these tasks is organized in the ChatML format, illustrated by the following example:
\begin{center}
\begin{tcolorbox}[colback=white, coltext=black, title=\textbf{Speech-to-text Data Format}]
\begin{verbatim}
{
    "messages": [
        {
            "role": "system",
            "content": "You are a helpful assistant."
        },
        {
            "role": "user",
            "content": "task_instruction <|startofspeech|> wav_path
            <|endofspeech|>"
        },
        {
            "role": "assistant",
            "content": "task_output"
        }
    ]
}
\end{verbatim}
\end{tcolorbox}
\end{center}
Here, ``\texttt{task\_instruction}'' corresponds to the natural language descriptions for different speech-to-text tasks. For instance, ``\texttt{Speech Transcription}'' may be used for speech recognition tasks, while ``\texttt{Translate \{SRC\_LANG\} into \{TGT\_LANG\}}'' may be used for speech translation tasks. ``\texttt{wav\_path}'' refers to the input audio file path, while ``\texttt{task\_output}'' refers to the output of each task.

\textbf{Text-to-Speech tasks.}  The data for this category mainly consists of basic speech synthesis data, which is the same data used for training CosyVoice 2. It includes 170,000 hours of text-speech paired data and supports four languages: Chinese, English, Korean, and Japanese. Additionally, there are approximately 1,000 hours of audio generation data controlled by instructions. The instructions are expanded to include natural language descriptions, generated by Qwen-Max\footnote{\url{https://help.aliyun.com/zh/model-studio/developer-reference/use-qwen-by-calling-api}}, utilizing human-labeled attributes such as emotion, speaking rate, dialect, and role-playing.

\begin{table}[htbp]
    \footnotesize
    \centering
    \scalebox{1.0}{
    \begin{tabularx}{\textwidth}{>{\raggedright\arraybackslash}X}
        \toprule
        \textbf{User:} Please speaking very fast: Today is a happy day, full of laughter and joy. \\
        \textbf{Assistant:} \faComment~~(Fast speaking rate) Today is a happy day, full of laughter and joy. \\
        \textbf{User:} Speaking with a tone of sadness: I miss my dear friend who moved away last month.  \\
        \textbf{Assistant:} \faComment~~(Sad) I miss my dear friend who moved away last month.  \\
        \bottomrule
    \end{tabularx}
    }
    \vspace{3mm}
    \caption{Examples of text-to-speech data controlled by instructions.}
    \label{tab:example_instruct}
\end{table}

\textbf{Speech-to-Speech tasks.} 
The Speech-to-Speech data is primarily sourced through simulation, encompassing approximately 10,000 hours of multi-turn conversational speech and 100 hours of style-controllable multi-turn conversational speech. The method for simulating speech-to-speech chat data is as follows:
\begin{itemize}[leftmargin=*,noitemsep]
    \item For text chat data primarily sourced from Alpaca~\citep{alpaca} and ShareGPT\footnote{\url{https://sharegpt.com/}}, we utilize the zero-shot in-context generation method from CosyVoice~\citep{DBLP:journals/corr/abs-2407-05407} to convert user text into user speech. We fine-tune CosyVoice's base model with 2 hours of data from a selected speaker to create a speech synthesis model for the target speaker, referred to as \textit{CosyVoice-SFT}. This model synthesizes the assistant's speech (i.e., system speech). The advantage of using zero-shot in-context generation for user speech synthesis is its ability to ensure diversity in the generated user speech, thereby enhancing the generalizability of MinMo.
    \item To address the differences between synthesized and real audio, we select suitable real speech from the ASR data as user speech queries and use the corresponding text as input for Qwen-Max to generate response text, which is then synthesized into assistant speech using the CosyVoice-SFT model. This approach further enhances the model's robustness to real user audio inputs.
    \item To generate conversational speech that covers different speaking styles, we initially employ Qwen-Max to create a rich collection of style-controllable, multi-turn text dialogues. User queries are converted into speech using zero-shot generation by Cosyvoice. Subsequently, we employ Cosyvoice 2 to generate the assistant's expressive speech. Specifically, we input the assistant's response content along with an instructional prompt into Cosyvoice 2 to synthesize speech in specific styles. Additionally, a small, diverse, and preliminary recorded voice corpus is used as prompt speech to synthesize the expressive response speech by zero-shot generation. The former method enhances the diversity of the simulated speech, while the latter more effectively builds the expressiveness of various styles.
\end{itemize}

\textbf{Speech-to-ControlToken task.} The Speech-to-ControlToken data primarily consists of two parts. The first part is extracted from existing \textit{real} voice interaction data, while the second part is \textit{simulated} using text dialogue data. Specifically, the existing real voice interaction data includes resources such as Alimeeting~\citep{yu2022m2met}, Fisher~\citep{cieri2004fisher}, and our in-house voice interaction data, a total of approximately 3000 hours. The simulated data mainly includes the open-source MOSS dataset~\citep{sun2024moss} and spoken dialogues by synthesizing our in-house text dialogue data, yielding about 1000 hours of voice chat data. When constructing duplex training data using these voice interaction data, we apply heuristic rules for automatically annotating duplex labels on the samples, as follows.

\begin{itemize}[leftmargin=*,noitemsep]
\item For assistant's turn-taking, the endpoint of the user’s turn is taken as the starting point of the assistant's turn. 
\item For user's turn-taking, a time gap T after the assistant’s turn ends
is taken as the starting point of the user's turn, where $T \sim \mathcal{N}(0.6, 0.4^2) $.
\item For user's back-channel, we select instances from the voice interaction data when the user (taking one speaker in a dialogue as the user) is unable to interrupt the other speaker and treat them as training samples of user's back-channels.
\end{itemize}

\subsection{Model Training}
\label{sec:minmo_training}

MinMo is trained progressively through four stages of alignment:  (1) Speech-to-Text Alignment, (2) Text-to-Speech Alignment, (3) Speech-to-Speech Alignment, and (4) Duplex Interaction Alignment. Through the four alignment stages, MinMo gains its end-to-end audio comprehension and generation capabilities while retaining the capabilities of the backbone text LLM, achieving low latency and facilitating a seamless voice chat experience for the user, similar to GPT-4o. The four stages are detailed as follows.

\begin{figure*}[t!]
  \centering
  \includegraphics[width=\linewidth]{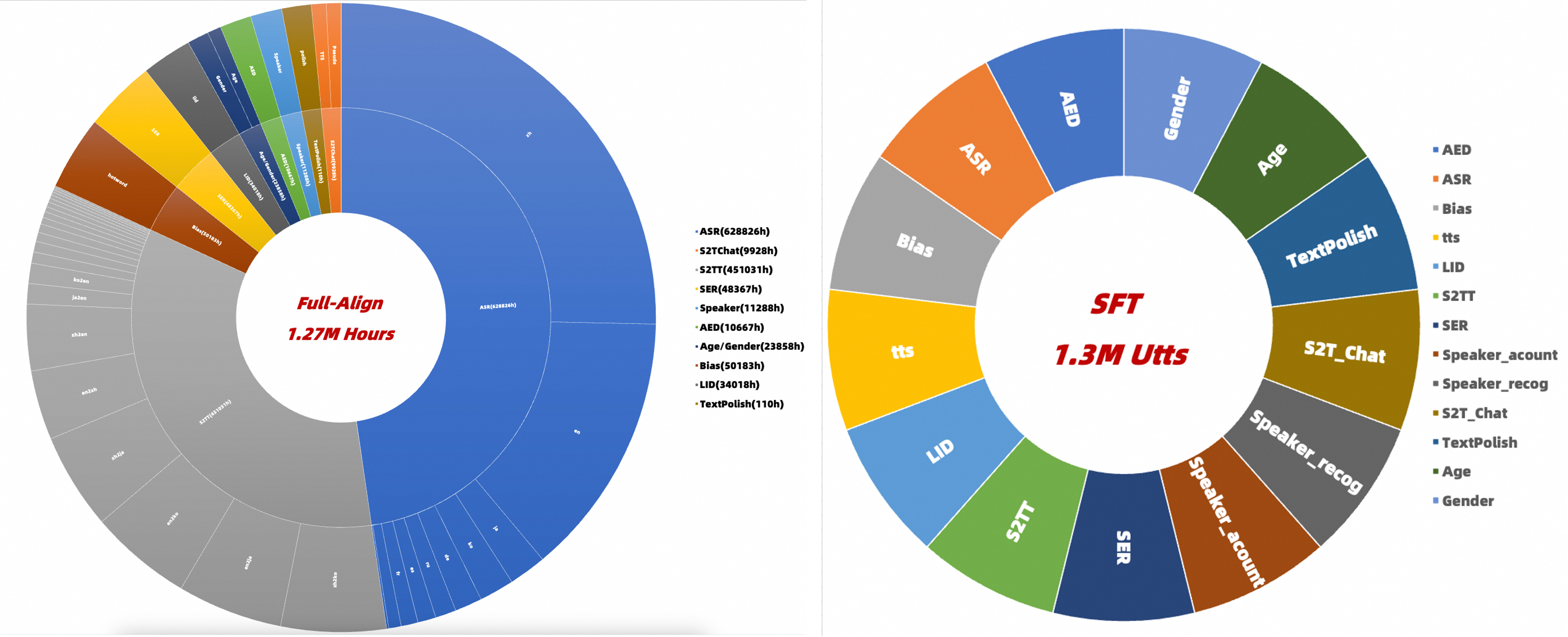}
  \caption{Detailed training data for the Speech-to-Text Alignment stage. \textbf{Left:} Data distribution for \textit{Full-Align} training. \textbf{Right:} Data distribution for instruction fine-tuning (SFT).}
  \label{fig:S2T_Data}
\end{figure*}

\textbf{Speech-to-Text Alignment.} This first stage aligns the audio modality's input latent space and the semantic space of a pre-trained text LLM using Speech-to-Text data shown in Table \ref{tab:MinMo_data}. This phase includes stepwise updates of the Input Projector and Voice Encoder in Figure~\ref{fig:MinMo}, as well as updating the text LLM using LoRA. Considering that the Voice Encoder and LLM (Qwen2.5-7B) are pre-trained while the Input Projector's parameters are randomly initialized, we perform a pre-alignment training (\textbf{Pre-align}) using a subset of the Speech-to-Text data shown in Table~\ref{tab:MinMo_data}, updating only the Input Projector. This Pre-align phase effectively prevents the randomly initialized parameters from having large-gradient influences on the pre-trained Voice Encoder at the initial training stage. After Pre-align, we use the full Speech-to-Text data for training both the Input Projector and the Voice Encoder while keeping LLM parameters frozen—a process called \textbf{Full-Align}. Following Full-Align, \textbf{instruction fine-tuning (SFT)} is conducted using approximately 1.3 million samples covering various tasks. During this stage, LLM is updated using LoRA, enhancing the model's ability to follow instructions. The specific data proportions used in the Full-Align and SFT stages are illustrated in Figure~\ref{fig:S2T_Data}. The Pre-Align phase uses about 1/10 of the Full-Align data.

\textbf{Text-to-Speech Alignment.} This second stage aligns the semantic space of the text LLM with the audio modality's output latent space, using Text-to-Speech data. This phase first trains the Output Projector and then jointly trains the Output Projector and the Voice Token LM while keeping other MinMo parameters frozen. In addition to the basic text-to-speech functionality, we leverage the end-to-end framework to enable MinMo to follow user instructions in voice interactions, delivering more expressive and entertaining audio responses. For instance, user can control the emotion, speaking rate, dialectal accent, or speaker style of the audio output via instructions. Approximately 1,000 hours of Instruct Speech Synthesis data are constructed, formatted as shown in Table~\ref{tab:example_instruct}.

\textbf{Speech-to-Speech Alignment.} This third stage continues training of MinMo using about 10,000 hours of paired audio data. Consistent with the Text-to-Speech Alignment phase, we continue updating only the Output Projector and the Voice Token LM. The training data for speech-to-speech alignment includes not only general speech-to-speech dialogues but also audio generation instructions with various settings, such as adopting specific dialects, speaking rates, and emotions for spoken dialogues. We find that even without updating the LLM, just by leveraging embeddings aligned with a small-scale instruction dataset (<150 hours), the large model can still learn quite effective control capabilities for audio generation. 

\textbf{Duplex Interaction Alignment.} After completing the first three training stages, MinMo acquires capabilities for audio understanding, audio generation, and half-duplex voice conversation. On this foundation, we further add a full-duplex module trained with 4,000 hours of long-form human-human spoken conversation. The Full Duplex Predictor module is exclusively trained during this stage. The Full Duplex Predictor takes the hidden embeddings of the LLM as input to predict whether the model needs to generate a response. The Full Duplex Predictor leverages the LLM's inherent semantic understanding capabilities to determine: 1) whether the model should respond to the current user query, and 2) whether the model should stop ongoing audio output to listen to the user query and provide an appropriate response.

\section{Experiments}
\label{sec:experiments}

We evaluate MinMo across multiple benchmarks, as detailed in Table \ref{tab:benchmark_summary}. These evaluation benchmarks cover speech recognition and speech translation tasks (multilingual speech recognition, multilingual speech translation, language identification, and contextual biasing speech recognition), speech analysis and understanding tasks (speech emotion recognition, speaker analysis, and audio event understanding), and speech-to-text enhancement tasks (spoken language smoothing, punctuation, and inverse text normalization). Additionally, we evaluate MinMo on voice generation tasks (text-to-speech and instruction-following voice generation) and voice chat tasks (including spoken question answering, spoken dialogue, and full-duplex interaction tasks).

\begin{table}[htbp]
\centering
\begin{adjustbox}{max width=\textwidth}
\begin{tabular}{lll}
\toprule
\textbf{Description} & \textbf{Dataset}   & \textbf{Metric} \\ \midrule
\multicolumn{3}{c}{\textbf{Speech Recognition and Translation}} \\\midrule
                                  &  Aishell-2~\citep{du2018aishell} & \multirow{5}{*}{CER\&WER} \\
                                  &  Wenetspeech~\citep{zhang2022wenetspeech}&  \\
Multilingual Speech Recognition      &  Librispeech~\citep{panayotov2015librispeech} &  \\
                            &  Fleurs~\citep{conneau2023fleurs}      &  \\
                                  &  CommonVoice~\citep{ardila2019common}  &     \\\midrule
Multilingual Speech Translation       &  Fleurs~\citep{conneau2023fleurs}   & \multirow{2}{*}{BLEU}\\
                           &  CoVoST2~\citep{wang2021covost} &   \\\midrule
Language Identification  & Fleurs~\citep{conneau2023fleurs} & Accuracy \\\midrule
Contextual Biasing Speech Recognition      &  Aishell-1-NE  & CER\&P\&R\\ \midrule
\multicolumn{3}{c}{\textbf{Speech Analysis and Understanding}} \\\midrule 

Speech Emotion Recognition        &  EMOBox~\citep{emobox} & UA\&WA\&F1\\\midrule
Audio Event Understanding   &  \multirow{4}{*}{AirBench\citep{yang2024air}} & \multirow{4}{*}{Accuracy}\\ 
Speaker Analysis for Gender and Age & \\
Vocal Sound                       & \\
Sound Question    &     & \\
 \midrule
\multicolumn{3}{c}{\textbf{Speech-to-Text Enhancement}} \\\midrule
Spoken Language Smoothing & SWAB~\citep{liu2024recording} & S-Faithful\&S-Formal \\ \midrule
Punctuation\&ITN & In-house testset & ChatGPT Score \\  \midrule
\multicolumn{3}{c}{\textbf{Voice Generation}} \\\midrule
\multirow{1}{*}{Text-to-Speech}          & Seed-TTS \citep{seed-tts}   & \multirow{1}{*}{WER\&CER\&NMOS} \\
\midrule
{Instruction-following Voice Generation}       & In-house testset  &{Accuracy} \\ \midrule
\multicolumn{3}{c}{\textbf{Voice Chat}} \\ \midrule
\multirow{3}{*}{Spoken Question Answering} & Web Questions \citep{DBLP:conf/emnlp/BerantCFL13} & \multirow{3}{*}{Accuracy} \\
                                           & Llama Questions \citep{nachmani2024spokenquestionansweringspeech} &  \\
                                           & TriviaQA \citep{joshi2017triviaqalargescaledistantly}   &     \\ \midrule
\multirow{2}{*}{Spoken Dialogue}           &  AlpacaEval \citep{alpaca_eval}   & \multirow{2}{*}{ChatGPT Score} \\
                                           & In-house ChitChat & \\\midrule

\multirow{3}{*}{Full Duplex} & Alimeeting \citep{yu2022m2met} & \multirow{3}{*}{Positive F1-Score} \\
                                           & Fisher \citep{cieri2004fisher} &  \\
                                           & Simulation \citep{sun2024moss}   &     \\
\bottomrule
\end{tabular}
\end{adjustbox}
\vspace{3mm}
\caption{Summary of evaluation benchmarks for MinMo in this report.}
\label{tab:benchmark_summary}
\end{table}

\subsection{Speech Recognition and Translation}
\paragraph{Multilingual Speech Recognition} We evaluate MinMo's speech-to-text transcription capabilities on public test sets in Mandarin, English, Japanese, Korean, and six other languages. These include Aishell-2 \citep{du2018aishell}, LibriSpeech test clean/other \citep{panayotov2015librispeech}, WenetSpeech \citep{zhang2022wenetspeech}, Fleurs \citep{conneau2023fleurs}, and Common Voice \citep{ardila2019common}. Table~\ref{tab:multilingual_asr_performance} presents the results from different models. For Mandarin (ZH), Japanese (JA), Korean (KO), and Cantonese (YUE), we employ the character error rate (CER) for evaluating transcription performance. For English (EN), German (DE), French (FR), Russian (RU), Spanish (ES), and Italian (IT), the word error rate (WER) is utilized as the evaluation metric. Note that all baseline model results are reproduced and processed using the same procedures as conducted on MinMo's results, for fair comparisons. Post-processing is based on a modified Whisper normalizer~\citep{radford2023robust}, with modifications primarily for improved number normalization. In Table~\ref{tab:multilingual_asr_performance}, the ``w/ LID'' column indicates that language identification (LID) information, such as English, Chinese, or Korean, is included as part of the decoding prompt, while the ``w/o LID'' column denotes results without the additional LID information. Results in parentheses are directly cited from the papers.

\begin{table}[htbp]
    \centering
    \begin{adjustbox}{max width=\textwidth}
    \begin{tabular}{lccccccc}
    \toprule
    \multirow{2}{*}{\textbf{Test set}} & \multirow{2}{*}{\textbf{Language}} & \multicolumn{2}{c}{\textbf{Whisper Large-v3}} & \multicolumn{2}{c}{\textbf{Qwen2-Audio}} & \multicolumn{2}{c}{\textbf{MinMo}} \\
       &  & \multicolumn{1}{c}{\textbf{w/ LID}} & \multicolumn{1}{c}{\textbf{w/o LID}} & \multicolumn{1}{c}{\textbf{w/ LID}} & \multicolumn{1}{c}{\textbf{w/o LID}} & \multicolumn{1}{c}{\textbf{w/ LID}} & \multicolumn{1}{c}{\textbf{w/o LID}} \\
     \midrule
     Aishell-2 Android & ZH & 5.14 & 4.96 & 3.02 (2.9) & 2.90 & 2.88 & \textbf{2.86}\\
     Aishell-2 iOS & ZH & 4.83 & 4.76 & 3.17 (3.0) & 3.06 & 2.73 & \textbf{2.69} \\
     Aishell-2 Mic & ZH & 4.98 & 4.89 & 3.22 (3.0) & 3.15 & \textbf{2.88} & 2.91 \\
     Wenetspeech test-net & ZH & 9.72 & 9.68 & 8.14 & 7.65 & 6.78 & \textbf{6.64}\\
     Wenetspeech test-meeting & ZH & 18.72 & 18.54 & 9.49 & 8.34 & \textbf{7.44} & 7.60\\
     Librispeech test-clean & EN & 2.56 & 1.90 & 1.75 (\textbf{1.6}) & 1.74 & 1.74 & 1.64\\
     Librispeech test-other & EN & 4.34 & 3.65 & 4.03 (\textbf{3.6}) & 4.13 & 3.89 & 3.82\\
     \midrule
     \multirow{10}{*}{Fleurs} & ZH & 4.65 & 4.75 & 3.80(7.5) & 4.14 & \textbf{2.95} & 3.38\\
                              & EN & 4.11 & 4.28 & 5.12 & 5.03 & 3.79 & \textbf{3.70} \\
                              & JA & 4.23 & 4.47 & 10.43 & 11.85 & \textbf{3.84} & 3.86\\
                              & KO & 3.33 & 3.20 & 10.57 & 22.01 & 2.92 & \textbf{2.85}\\
                              & YUE & 7.67 & 7.11 & 4.09 & \textbf{4.07} & 4.25 & 4.22\\
                              & DE & \textbf{5.05} & 5.09 & 10.48 & 12.77 & 5.22 & 5.18\\
                              & FR & 5.27 & 5.16 & 9.36 & 11.01 & \textbf{5.15} & 5.33\\
                              & RU & \textbf{5.05} & 5.12 & 23.2 & 40.4 & 6.23 & 6.18\\
                              & ES & 2.9 & \textbf{2.86} & 7.31 & 18.02 & 3.44 & 3.55\\
                              & IT & 2.5 & \textbf{2.41} & 6.74 & 8.42 & 3.48 & 3.34\\
                & \textbf{Avg.} & 4.48 & 4.45 & 9.11 & 13.77 & \textbf{4.13} & 4.16 \\
    \midrule
    \multirow{10}{*}{Common Voice} & ZH & 12.4 & 12.52 & 6.48 (6.9) & 18.83 & 6.34 & \textbf{6.31}\\
                                   & EN & 9.66 & 17.91 & 8.82 (8.6) & 9.15 & \textbf{7.92} & 8.16\\
                                   & JA & \textbf{10.30} & 10.32 & 13.51 & 13.58 & 13.41 & 11.04\\
                                   & KO & 11.74 & \textbf{5.03} & 17.53 & 19.37 & 6.61 & 6.33\\
                                  & YUE & 10.25 & 37.71 & 6.01 (\textbf{5.9}) & 6.09 & 6.35 & 6.30\\
                                   & DE & \textbf{5.95} & 6.41 & 7.63 & 7.93 & 6.56 & 6.61\\
                                   & FR & 11.22 & 11.47 & 9.63 (9.6) & 11.10 & \textbf{8.46} & 8.59\\
                                   & RU & \textbf{5.94} & 6.48 & 16.82 & 22.87 & 6.97 & 7.12\\
                                   & ES & \textbf{4.94} & 5.19 & 5.72 & 9.14 & 4.96 & 4.99\\
                                   & IT & \textbf{5.77} & 6.31 & 6.81 & 7.96 & 6.08 & 6.16\\
                        & \textbf{Avg.} & 8.82 & 11.94 & 9.90 & 12.60 & 7.37 & \textbf{7.16}\\
    \bottomrule
    \end{tabular}
    \end{adjustbox}
    \vspace{3mm}
    \caption{Multilingual speech recognition results from our MinMo and baseline models in terms of word error rate (WER) and character error rate (CER) on Mandarin, English, and multilingual public test sets. Results in parentheses are directly cited from papers. The best result for each test set is boldfaced.}
    \label{tab:multilingual_asr_performance}
\end{table}

\textbf{As shown in Table \ref{tab:multilingual_asr_performance}, MinMo achieves superior ASR performance on most test sets across various languages, compared to Whisper Large v3~\citep{radford2023robust} and Qwen2-Audio~\citep{chu2024qwen2}}. 
The ``w/ LID'' columns for Whisper Large-v3 and Qwen2-Audio show similar results to those reported in the original papers. Testing on Common Voice with or without LID information as a prompt shows a significant gap in average error rates for Whisper Large v3 and Qwen2-Audio, indicating that these two models strongly depend on the LID information. In contrast, \textbf{MinMo demonstrates robust and consistent ASR performance regardless of the presence of the language identification.}

\paragraph{Multilingual Speech Translation} 
We evaluate speech-to-text translation capabilities on the Fleurs \citep{conneau2023fleurs} and CoVoST2 \citep{wang2021covost} test sets. On the Fleurs test set, we report results for all translation directions supported by our model; whereas, on the CoVoST2 test set, we only report results for translating from English to other languages (en2xx) and vice versa (xx2en), due to dataset limitations, as it primarily focuses on English-centric translation pairs. As shown in Table~\ref{tab:s2tt_performance}, our end-to-end MinMo consistently outperforms the cascaded model by pipelining Whisper Large V3 and Qwen2.5-7B-Instruct, in terms of BLEU scores. 
\textbf{Compared to other end-to-end baselines, MinMo achieves SOTA performance on Chinese$\leftrightarrow$English and Japanese$\leftrightarrow$English translations and top-tier performance on other language pairs}. We attribute this strong performance to the extensive speech translation training data (451K hours of S2TT training data as in Table~\ref{tab:MinMo_data}) and the powerful audio encoder. 
Notably, even though we only augment our training data with the CoVoST2 set, excluding the Fleurs set, our model maintains consistent performance across both test sets, indicating high robustness.

\begin{table}[htbp]
    \centering
    \begin{adjustbox}{max width=\textwidth}
    \begin{tabular}{lccccc}
    \toprule
    \textbf{Test set} & \makecell{\textbf{Language} \\ \textbf{Directions}} & \textbf{Qwen2-Audio}\textsuperscript{†} & \makecell{\textbf{SeamlessM4T} \\ \textbf{Large v2}} & \makecell{\textbf{Whisper large-v3} \\ \textbf{+Qwen2.5-7B-Instruct}} & \textbf{MinMo} \\ \midrule
    \multirow{26}{*}{Fleurs}& zh2en & 20.50 & 22.98 & 22.98 & \textbf{24.71} 
\\
    & ja2en & 2.94 & 18.23 & 20.76 & \textbf{24.09} 
\\
    & ko2en & 6.73 & 24.11 & 24.01 & \textbf{25.44} 
\\
    & yue2en & 16.16 & 19.04 & 21.18 & \textbf{23.87} 
\\
    & de2en & 30.58 & 37.00 & 37.18 & \textbf{39.54} 
\\
    & fr2en & 29.56 & 33.97 & 35.46 & \textbf{36.5} 
\\
    & ru2en & 20.09 & 30.17 & 31.57 & \textbf{32.71} 
\\
    & it2en & 21.86 & 26.50 & 27.01 & \textbf{27.61} 
\\
    & es2en & 21.93 & \textbf{31.55} & 26.80 & 27.67 
\\
 & \textbf{xx2en avg}& 18.93 & 27.06 & 27.44 & \textbf{29.13} 
\\ \cmidrule(lr){2-6}& en2zh & 30.46 & 29.81 & 38.83 & \textbf{40.68} 
\\
    & ja2zh & 7.46 & -& \textbf{25.57} & 25.14 
\\
    & ko2zh & 11.63 & -& \textbf{29.04} & 23.39 
\\
    & yue2zh & 29.20 & -& 32.52 & \textbf{35.26} 
\\
    & de2zh & 28.49 & -& 33.38 & \textbf{35.22} 
\\
    & fr2zh & 29.03 & -& 32.38 & \textbf{34.31} 
\\
    & ru2zh & 21.48 & -& 31.96 & \textbf{34.34} 
\\
    & it2zh & 26.64 & -& 30.37 & \textbf{33.30} 
\\
    & es2zh & 25.53 & -& 30.32 & \textbf{32.05} 
\\
 & \textbf{xx2zh avg}& 23.32 & -& 31.60 & \textbf{32.63} 
\\ \cmidrule(lr){2-6}& zh2ja & 16.29 & -& 17.36 & \textbf{19.49} 
\\
    & en2ja & 21.10 & \textbf{35.24} & 23.39 & 30.26 
\\
 & \textbf{xx2ja avg}& 18.70 & -& 20.38 & \textbf{24.88} 
\\ \cmidrule(lr){2-6}& zh2ko & 8.39 & -& 9.97 & \textbf{17.53} 
\\
    & en2ko & 10.74 & 12.82 & 12.42 & \textbf{26.06} 
\\
 & \textbf{xx2ko avg}& 9.57 & -& 11.20 & \textbf{21.80} 
\\ \midrule
    \multirow{10}{*}{CoVoST2}& en2zh & 45.20 & 35.90 & 38.86 & \textbf{46.68} 
\\
    & en2ja & 28.76 & \textbf{39.70} & 22.04 & 35.10 
\\
    & zh2en & 24.40 & 22.18 & 20.56 & \textbf{25.95} 
\\
    & ja2en & 20.68 & 23.79 & 25.49 & \textbf{28.87} 
\\
    & de2en & 35.20 & \textbf{39.94} & 35.10 & 39.91 
\\
    & fr2en & 38.50 & \textbf{42.10} & 35.95 & 41.31 
\\
    & ru2en & 40.75 & \textbf{53.58} & 42.58 & 48.60 
\\
    & it2en & 36.30 & 39.97 & 30.80 & \textbf{40.62} 
\\
    & es2en & 40.00 & 42.94 & 38.57 & \textbf{43.26} 
\\
    \cmidrule(lr){2-6}& \textbf{CoVoST2 avg} & 34.42 & 37.79 & 32.22 & \textbf{38.92} \\ \bottomrule
    \end{tabular}
    \end{adjustbox}
    \vspace{3mm}
    \caption{Multilingual speech translation results from MinMo and baseline models on CoVoST2 and Fleurs test sets, in terms of BLEU score (higher is better). Results marked with \textsuperscript{†} are obtained using translation prompts that specify the source language, rather than only indicating the target language. Results not reported in the original papers have been reproduced by the authors of this work.}
    \label{tab:s2tt_performance}
\end{table}

\paragraph{Language Identification}
For evaluation of language identification performance, we use the Fleurs dataset, which covers 102 languages. \textbf{MinMo achieves language identification accuracy of 85.3\%, outperforming all previous models shown in Table~\ref{tab:language_identification}}. Specifically, zero-shot Whisper-V3 often miscategorizes Cantonese as Chinese while MinMo accurately identifies Cantonese. The accuracy of Zero-shot Whisper is not competitive due to the fact that it misses training data for 20 languages in the Fleurs dataset; whereas, the LID training data of MinMo covers all 102 languages in Fleurs.

\begin{table}[htbp]
\centering
\begin{tabular}{lc}
\toprule
\multicolumn{2}{c}{Fleurs(102)}\\ \midrule
w2v-bert-51 (0.6B) & 71.4        \\
mSLAM-CTC (2B)     & 77.7      \\
Zero-shot Whisper  & 64.5     \\
\midrule
MinMo              & \textbf{85.3}    \\ \bottomrule
\end{tabular}
\vspace{3mm}
\caption{Language identification results from MinMo and baseline models on the Fleurs data set (covering 102 languages), in terms of Accuracy.}
\label{tab:language_identification}
\end{table}

\paragraph{Contextual Biasing Speech Recognition}
Contextual biasing, or hotword customization, allows users to obtain customized ASR results with specific contexts or hotwords. MinMo enhances ASR capabilities by integrating advanced prompts for contextual biasing. We prepare corresponding training data for alignment and SFT stages, by organizing hotwords within prompts preceding speech processing instructions, to enable effective customization. Evaluations include hotwords biasing test and general biasing test, as shown in Table~\ref{hwtest}. The hotwords biasing test involves three data sets used by SeACo-Paraformer~\citep{10446106}, which contain hotwords for biasing evaluation. The general biasing test uses data sets with fewer hotwords to assess resistance to irrelevant ones.

\begin{table}[htbp]
\centering
\begin{adjustbox}{max width=0.85\textwidth}
\begin{tabular}{llccc}
\toprule
\textbf{}                                       & \textbf{Test Name} & \textbf{utt.} & \textbf{hotwords prop.} & \textbf{hotwords(hc)} \\ \toprule
\multirow{3}{*}{\textbf{Hotwords Biasing Test}} & Test-Commercial    & 2000          & 100\%                   & 693(72)               \\
                                                & Dev-Aishell1-NE    & 1334          & 100\%                   & 600(371)               \\
                                                & Test-Aishell1-NE   & 808           & 100\%                   & 400(226)               \\ \midrule
\multirow{2}{*}{\textbf{General Biasing Test}}  & Cor-Chinese        & 1749          & 5.2\%                   & 87                \\
                                                & Cor-English        & 1211          & 28.8\%                  & 106               \\ \bottomrule
\end{tabular}
\end{adjustbox}
\vspace{3mm}
\caption{Statistics of the test sets for evaluating contextual biasing speech recognition. (\textbf{hc} denotes hard-case hotwords, which are hotwords with recall rates under 40\% in basic speech recognition.}
\label{hwtest}
\end{table}

We compare MinMo with the baseline model SeACo-Paraformer~\citep{10446106}, which is an open-source ASR system with strong Chinese hotword biasing performance. Table~\ref{hwtest1} shows that \textbf{MinMo outperforms the competitive baseline SeACo-Paraformer in terms of ASR accuracy (both with and without hotwords) and also recall rates of hard-case hotwords}. Table~\ref{hwtest2} further demonstrates that \textbf{MinMo achieves contextual biasing capability in multiple languages, without compromising its general ASR performance}.

\begin{table}[htbp]
\centering
\begin{adjustbox}{max width=\textwidth}
\begin{tabular}{lcccccccccccc}
\toprule
\textbf{}        & \multicolumn{6}{c}{\textbf{SeACo-Paraformer†}}                                                                                                        & \multicolumn{6}{c}{\textbf{MinMo}}                                                                                                                    \\ \cmidrule{2-13} 
\textbf{}        & \multicolumn{3}{c}{\textbf{w/o bias}}                                     & \multicolumn{3}{c}{\textbf{w/ bias}}                                      & \multicolumn{3}{c}{\textbf{w/o bias}}                                     & \multicolumn{3}{c}{\textbf{w/ bias}}                                      \\ \cmidrule{2-13} 
\textbf{}        & \textbf{CER$\downarrow$} & \textbf{R$\uparrow$} & \textbf{R-hc$\uparrow$} & \textbf{CER$\downarrow$} & \textbf{R$\uparrow$} & \textbf{R-hc$\uparrow$} & \textbf{CER$\downarrow$} & \textbf{R$\uparrow$} & \textbf{R-hc$\uparrow$} & \textbf{CER$\downarrow$} & \textbf{R$\uparrow$} & \textbf{R-hc$\uparrow$} \\ \midrule
Test Commercial  & 3.65                     & 90.0                 & 11.0                    & 3.10                     & 97.0                 & 68.0                    & \textbf{2.63}            & 92.3                 & 58.3                    & \textbf{2.45}            & 97.1                 & \textbf{84.9}           \\
Dev-Aishell1-NE  & 5.47                     & 54.0                 & 3.0                     & 1.98                     & 94.0                 & 84.0                    & \textbf{2.32}            & 70.8                 & 56.5                    & \textbf{1.16}            & 93.5                 & \textbf{90.5}           \\
Test-Aishell1-NE & 5.55                     & 56.0                 & 3.0                     & 2.27                     & 94.0                 & 87.0                    & \textbf{2.41}            & 68.3                 & 49.3                    & \textbf{1.19}            & 91.1                 & 86.4           \\ \bottomrule
\end{tabular}
\end{adjustbox}
\vspace{3mm}
\caption{Performance of contextual biasing speech recognition from MinMo and the baseline model on the hotwords biasing test sets, in terms of character error rate (CER), recall of hotwords (R), and recall of hard-case hotwords (R-hc).}
\label{hwtest1}
\end{table}

\begin{table}[htbp]
\centering
\begin{adjustbox}{max width=0.59\textwidth}
\begin{tabular}{lcccccc}
\toprule
\textbf{}       & \multicolumn{3}{c}{\textbf{w/o bias}}                      & \multicolumn{3}{c}{\textbf{w/ bias}}                         \\ \cmidrule{2-7} 
\textbf{}       & \textbf{CER$\downarrow$} & \textbf{P$\uparrow$}           & \textbf{R$\uparrow$}           & \textbf{CER$\downarrow$}   & \textbf{P$\uparrow$}           & \textbf{R$\uparrow$}           \\ \toprule
Correct-Chinese & 12.92        & 89.2                 & 15.9                 & \textbf{12.64}          & 92.1                 & \textbf{69.0}                 \\
Correct-English & 18.68        & 85.3                 & 46.2                 & \textbf{18.48} & 87.4                 & \textbf{53.4}                 \\ \bottomrule
\end{tabular}
\end{adjustbox}
\vspace{3mm}
\caption{Performance of contextual biasing speech recognition from MinMo on the general biasing test sets, in terms of character error rate (CER), precision (P), and recall (R).}
\label{hwtest2}
\end{table}

\subsection{Speech Analysis and Understanding}

\paragraph{Speech Emotion Recognition}

We evaluate the Speech Emotion Recognition (SER) capability of MinMo using seven widely used emotion recognition datasets from EmoBox, including CREMA-D~\citep{cremad}, MELD~\citep{meld}, IEMOCAP~\citep{IEMOCAPIE}, MSP-Podcast~\citep{msppodcast}, CASIA~\citep{casia}, MER2023~\citep{lian2023mer2023multilabellearning}, and ESD~\citep{esd}. These datasets include both Chinese and English languages and scenarios such as acting, TV dramas, and daily conversations. We adopt unweighted average accuracy (UA), weighted average accuracy (WA), and macro F1 Score (F1) as evaluation metrics. Results on these test sets from the recent SER toolkit EmoBox~\citep{emobox} are cited.
We also evaluate the baseline audio-LLM models SALMONN and Qwen-Audio using their released model checkpoints. As shown in Table~\ref{tab:ser_performace}, MinMo outperforms the baseline audio-LLM models on most datasets, particularly achieving nearly 100\% accuracy on acting audio datasets (CASIA, CREMA-D, ESD). It is important to point out that the comparison between MinMo and the baseline models SALMONN and Qwen-Audio on the aforementioned datasets is inconclusive, since the optimal prompts and post-processing methods for the baseline models are unclear. Therefore, we further utilize the Air-Bench benchmark, which is specifically designed for evaluating large audio language models with \textit{standardized} post-processing and scoring scripts for fair comparison. \textbf{As shown in Table~\ref{tab:result_airbench}, MinMo outperforms all the baseline models on all tasks on this benchmark, including Language ID, Gender, Age, Emotion, Vocal Sound classification tasks, except for being outperformed by Qwen-Audio on the sound question classification task}.

In addition to Chinese and English languages, we also evaluate MinMo on low-resource languages in a zero-shot setting, as shown in Table~\ref{tab:emobox}. All utterances and reference labels are derived from the EmoBox benchmark, which provides the official training and evaluation partitions for the 32 publicly available SER datasets in 14 languages. EmoBox evaluates the SER capabilities of 10 different pre-trained models across all datasets,
with the classification results used as reference labels in Table~\ref{tab:emobox}. Despite the absence of in-domain audio used in training, MinMo achieves the best F1 score on most languages, even for those languages not included in the original training data of MinMo. \textbf{These results highlight MinMo's excellent cross-lingual generalization capability}.

\begin{table}[htbp]
  \centering
  \resizebox{\textwidth}{!}{%
    \begin{tabular}{l  ccc  ccc  ccc  ccc}
      \toprule
      \multirow{2}{*}{\textbf{Test Set}}  & \multicolumn{3}{c}{\textbf{MinMo}} & \multicolumn{3}{c}{\textbf{SALMONN}}  & \multicolumn{3}{c}{\textbf{Qwen-Audio}} & \multicolumn{3}{c}{\textbf{EmoBox}} \\
       & UA & WA & F1 & UA & WA & F1 & UA & WA & F1 & UA & WA & F1 \\
      \midrule
                CASIA         &\textbf{98.1}&\textbf{98.1}&\textbf{98.1} &35.9&35.6&33.8     &40.0&38.4&36.0    & 59.6&59.6&56.3   \\
      CREMA-D     &\textbf{94.8}&\textbf{94.8}&\textbf{94.8} &49.7&50.2&43.7     &82.2&83.8&83.2     & 76.8 & 76.5 &76.6 \\
                ESD        &\textbf{99.9}&\textbf{99.9}&\textbf{99.9} &33.9&34.5&31.7     &47.3&47.4&43.6  &84.6&84.6&84.3      \\
      IEMOCAP    &\textbf{74.9}&\textbf{75.7}&\textbf{74.2} &59.0&60.6&59.0     &69.6&67.6&62.9      & 73.5 & 72.9 & 73.1\\
                MELD       &\textbf{61.0}&\textbf{65.1}&\textbf{54.0} &39.2&47.2&39.6     &49.9&56.8&46.8     & 31.5 & 51.9 & 32.9  \\
                MSPPodcast   &\textbf{66.4}&\textbf{74.5}&\textbf{62.5} &40.1&58.0&40.3   &57.9&70.0&54.6    & 21.4 & 43.4 & 21.5   \\
      \bottomrule
  \end{tabular}}
        \vspace{3mm}
  \caption{Speech emotion recognition performance from MinMo and the baseline models on various evaluation benchmarks, in terms of unweighted average accuracy (UA), weighted average accuracy (WA), and macro F1 score (F1). Results for SALMONN and Qwen-Audio are reproduced by the authors of this work.}
\label{tab:ser_performace}
\end{table}

\begin{table}[htbp]
    \centering
    \resizebox{\textwidth}{!}{%
    \begin{tabular}{l*{6}{c}}
        \toprule
        \multirow{2}{*}{\textbf{Task}} & \multirow{2}{*}{\textbf{MinMo}} & \textbf{Qwen-Audio} & \multirow{2}{*}{\textbf{Qwen-Audio}} & \multirow{2}{*}{\textbf{Pandagpt}} & \multirow{2}{*}{\textbf{SALMONN}} & \multirow{2}{*}{\textbf{Next-gpt}} \\
         &  & \textbf{Turbo} & & & & \\
        \midrule
        Language ID    & \textbf{99.2\%} & 95.9\% & 92.8\% & 34.6\% & 28.1\% & 23.7\%  \\
        Gender       & \textbf{86.7\%} & 82.5\% & 67.2\% & 66.5\% & 35.5\% & 57.0\% \\
        Age          & \textbf{70.1\%} & 58.8\% & 36.0\% & 42.5\% & 48.7\% & 62.4\% \\
        Emotion      & \textbf{64.5\%} & 60.0\% & 43.2\% & 26.0\% & 29.9\% & 25.7\% \\
        Vocal Sound  & \textbf{93.0\%} & 78.1\% & 84.9\% & 31.6\% & 45.3\% & 23.5\% \\
        Sound Question    & 50.3\% & 62.8\% & \textbf{64.6\%} & 48.7\% & 28.4\% & 18.8\% \\
        \bottomrule
    \end{tabular}
    }
    \vspace{3mm}
    \caption{Performance comparison between MinMo and the baseline models on the AIR-Bench benchmark, including Language ID, Gender, Age, Emotion, Vocal Sound, and Sound Question classification tasks, in terms of Accuracy. Results of Qwen-Audio and Qwen-Audio-Turbo~\citep{qwen-audio_2023}, Pandagpt~\citep{pandagpt}, SALMONN~\citep{salmonn} and Next-gpt~\citep{DBLP:conf/icml/Wu0Q0C24} are cited from the official AIR-Bench website.
    }
    \label{tab:result_airbench}
    
\end{table}

\begin{table}[htbp]
  \centering
  \resizebox{\textwidth}{!}{
    \begin{tabular}{l  ccc  ccc  ccc ccc}
      \toprule
      \multirow{2}{*}{Model} & UA & WA & WF1     & UA & WA & WF1      & UA & WA & WF1  & UA & WA & WF1     \\
      & \multicolumn{3}{c}{AESDD (el)}     & \multicolumn{3}{c}{CAFE (fr)}     & \multicolumn{3}{c}{ RESD (ru)}    & \multicolumn{3}{c}{  ASED (am)}               \\
      \midrule
      wavlm-large          & 78.9 & 67.6 & 67.6      & 62.2 & 61.3 & 61.1      & 55.8 & 56.4 & 55.8  & 96.4 & 96.4 & 96.4\\
      data2vec2.0-large    & 72.2 & 72.3 & 71.8      & 59.0 & 58.0 & 57.5      & 44.0 & 44.6 & 44.2  & 94.3 & 94.2 &  94.2\\
      whisper-large-v3     & 79.1 & 79.1 & 79.1      & 69.4 & 68.8 & 68.0      & 54.9 & 55.5 & 54.9  & \textbf{96.7} & \textbf{96.7} & \textbf{96.7} \\
      MinMo                & \textbf{96.1} & \textbf{96.5} & \textbf{96.2}         & \textbf{94.9} & \textbf{94.8} & \textbf{94.9}        & \textbf{76.7} & \textbf{78.6} & \textbf{76.3}  & 75.3 & 75.2 & 73.7\\
                \midrule
                Model &\multicolumn{3}{c}{EmoDB (de)}  &\multicolumn{3}{c}{EMOVO (it)} & \multicolumn{3}{c}{MESD (es)} & \multicolumn{3}{c}{ Polish (pl)}    \\
                \midrule
      wavlm-large          & 92.5 & 92.6 & 92.5      & 48.8 & 48.8 & 44.1      & 62.5 & 62.4 & 62.3  & 79.2 & 79.2 & 79.0\\
      data2vec2.0-large    & 79.3 & 80.4 & 79.9      & 45.8 & 45.8 & 43.6      & 48.4 & 48.4 & 46.8  & 74.0 & 74.0 & 74.0\\
      whisper-large-v3     & 91.2 & 92.4 & 91.8      & 57.8 & 57.8 & 56.0      & 69.7 & 69.6 & 69.6  & \textbf{83.2} & \textbf{83.2} & \textbf{82.7}\\
      MinMo                & \textbf{98.4} & \textbf{98.7} & \textbf{98.7}    & \textbf{81.7} & \textbf{81.1} & \textbf{68.3}        & \textbf{89.5} & \textbf{89.1} & \textbf{88.6}  & 58.1 & 58.1 & 55.9\\
                \midrule
                Model &\multicolumn{3}{c}{ SUBESCO (bn)}  &\multicolumn{3}{c}{ShEMO (fa)} & \multicolumn{3}{c}{ URDU (ur)}   & \multicolumn{3}{c}{  TurEV-DB (tr)}  \\
                \midrule
      wavlm-large          & 65.3 & 65.3 & 65.0      & 71.7 & 87.1 & 73.5      & \textbf{86.6} & \textbf{86.6} & \textbf{86.6}  & 79.5 & 80.0 & 79.5\\
      data2vec2.0-large    & 66.4 & 66.4 & 66.2      & 64.0 & 82.6 & 68.4      & 78.1 & 78.1 & 78.1  & 63.7 & 64.1 & 63.0\\
      whisper-large-v3     & 73.0 & 73.0 & 72.9      & \textbf{80.2} & \textbf{89.5} & \textbf{82.9}      & 82.5 & 82.5 & 82.4  & 81.3 & 81.5 & 81.3\\
      MinMo                & \textbf{80.1} & \textbf{80.5} & \textbf{72.8}    & 70.9 & 84.0 & 72.7        & 82.1 & 82.1 & 79.4  & \textbf{82.0} & \textbf{82.1} & \textbf{81.6}\\
      \bottomrule
  \end{tabular}}
        \vspace{3mm}
  \caption{SER performance of MinMo and the baseline models on the multi-languages datasets from EmoBox, in terms of unweighted average accuracy (UA), weighted average accuracy (WA), and macro F1 score (F1).}
  \label{tab:emobox}
\end{table}

\paragraph{Audio Event Understanding}
We compare MinMo's voice and audio event understanding capabilities against other Audio-LLM models, using the Air-Bench benchmark. The results are shown in Table~\ref{tab:result_airbench}. On the voice sound classification task (Vocal Sound), MinMo surpasses all baseline models.
However, we find that on more complex sound question-answering tasks, MinMo performs worse than Qwen-Audio although still outperforming other models.
This can be attributed to two factors: first, with the voice encoder and the training paradigm, MinMo is primarily designed for voice interaction, hence some sound questions may exceed its scope; second, during evaluation, MinMo predicts what happens in the audio rather than strictly choosing the options provided by the Air-Bench, hence some correct or similar-to-correct responses generated by MinMo are aligned with incorrect choices by the post processing script.

\paragraph{Speaker Analysis}
Speaker analysis involves several tasks that are essential for understanding and interacting with audio data, including gender detection, age estimation, speaker counting, speaker identification, multi-speaker recognition, and target speaker recognition. In this report, we focus on evaluating MinMo's performance in gender detection and age estimation.
Table~\ref{tab:result_airbench} compares MinMo against the baseline models on these tasks on the AIR-Bench benchmark, in terms of classification accuracy. \textbf{The results reveal that MinMo outperforms all the baseline models on gender detection and age estimation tasks}.

\subsection{Speech-to-Text Enhancement}

\paragraph{Spoken Language Smoothing}

The spoken language smoothing task takes the ASR transcripts of spoken language, and outputs formal-style written text. Examples of spoken language smoothing are shown in Table~\ref{tab:sample_smooth}. For this task, we construct a multi-domain dataset for training and evaluation, by extending the SWAB dataset~\citep{liu2024recording} that we create for spoken-to-written conversion of ASR transcripts. The SWAB dataset is derived from Chinese and English meetings, podcasts, and lectures. After the generation of ASR transcripts for the original videos and audios, approximately ten annotators create formal-style written text based on the ASR transcripts while preserving their original content.  The training set of SWAB comprises 20,000 paragraphs, and the test set includes 100 randomly sampled paragraphs in both Chinese and English. We conduct full fine-tuning and compare MinMo with Qwen2.5-7B-based model on the SWAB test set, with results shown in Table~\ref{tab:result_smooth}. For objective metrics, we calculate BLEU~\citep{DBLP:conf/acl/PapineniRWZ02}, ROUGE~\citep{lin2004rouge}, and BLEURT~\citep{DBLP:conf/acl/SellamDP20} with the human target as reference. Notably, we observe that the spoken language smoothing task shows significant subjectivity and diversity; therefore, objective metrics based on lexical matching may not adequately reflect the performance. Consequently, we use human and LLM annotations to provide rankings of \textbf{faithfulness (S-Faithful} (i.e., faithfulness to the original content) and \textbf{formality (S-Formal)}. The prompts for automated LLM scoring are presented in Appendix~\ref{appendix_spoken_Language_smoothing}. 
Table~\ref{tab:result_smooth} shows that \textbf{the performance of our model and Qwen2.5-7B is comparable, suggesting that MinMo possesses reasonable capability to smooth spoken language}.

\begin{table}[htbp]
    \centering
    \scalebox{0.8}{
        \begin{tabular}{ll}
            \toprule
            \textbf{Source} & eh? You have more 1, eh, more one voice, eh, eh, produced in the in the moment. \\
            \textbf{Target} & You have more voices producing in the moment. \\
            \midrule
            \textbf{Source} & so what, what do they need to do left?  \\
            \textbf{Target} & What tasks do they need to complete? \\
            \midrule
            \multirow{2}{*}{\textbf{Source}} & Well, I think you two, especially you and, and Daniel, you both had, the less creative, roles and \\
             & the project. That's true. Of course.  \\
            \textbf{Target} & I believe that you two, especially you and Daniel, had the less creative roles in the project. \\
            \bottomrule
        \end{tabular}
    }
    \vspace{3mm}
    \caption{Examples of the source and the prediction of MinMo for the spoken language smoothing task, sampled from the SWAB test set.}
    \label{tab:sample_smooth}
\end{table}

\begin{table}[htbp]
    \centering
    \scalebox{0.9}{
        \begin{tabular}{lcccccccc}
            \toprule
            \multirow{2}{*}{\textbf{Model}} & \multirow{2}{*}{\textbf{CER$\downarrow$}} & \multirow{2}{*}{\textbf{BLEU$\uparrow$}} & \multirow{2}{*}{\textbf{ROUGE-L$\uparrow$}} & \multirow{2}{*}{\textbf{BLEURT$\uparrow$}} & \multicolumn{2}{c}{\textbf{S-Faithful$\uparrow$}} & \multicolumn{2}{c}{\textbf{S-Formal$\uparrow$}} \\
            & & & & & Human & LLM & Human & LLM \\
            \midrule
            Qwen2.5-7B & 0.75 & 21.06 & \textbf{42.01} & \textbf{64.61} & \textbf{9.04} & \textbf{8.06} & \textbf{9.80} & \textbf{7.63} \\
            MinMo & \textbf{0.72} & \textbf{22.31} & 41.42 & 61.50 & 8.66 & 7.59 & 9.72 & 7.52 \\
            \bottomrule
        \end{tabular}
    }
    \vspace{3mm}
    \caption{Spoken language smoothing performance of MinMo and Qwen2.5-7B on the SWAB test set, in terms of objective metrics (CER, BLEU, ROUGE-L, and BLEURT)}
    \label{tab:result_smooth}
\end{table}

\paragraph{Punctuation Insertion and Inverse Text Normalization}

\begin{table}[htbp]
\centering
\begin{tabular}{lcccc}
\toprule
\multicolumn{1}{c}{\multirow{2}{*}{Model}} & \multicolumn{2}{c}{\textbf{PUNC}} & \multicolumn{2}{c}{\textbf{ITN}} \\ \cmidrule(lr){2-5} 
\multicolumn{1}{c}{}                        & Fleurs-zh          & Fleurs-en         & Fleurs-zh         & Fleurs-en         \\ \midrule
SenseVoice-L                                & 2.49           & 1.40          & 2.48       & 1.48       \\
whisper-large-v3                            & 1.23           & 2.49          & 1.39       & 2.43       \\
\toprule
MinMo                                       & \textbf{2.65}           & \textbf{2.58}          & \textbf{2.61}       & \textbf{2.57}       \\ \bottomrule
\end{tabular}
\vspace{3mm}
\caption{GPT-4 Turbo ranking scores of punctuation insertion and ITN for MinMo and the baseline models on the Chinese and English subsets of the Fleurs dataset.}
\label{tab:result_itn_punc}
\end{table}

For the punctuation insertion (PUNC) and Inverse Text Normalization (ITN) tasks, we use the Chinese and English data from the Fleurs dataset. We compare MinMo against SenseVoice-L and whisper-large-v3, as shown in Table~\ref{tab:result_itn_punc}. Given the subjectivity of the punctuation insertion and ITN tasks, we employ GPT-4 Turbo to rank the three outcomes for evaluation. The task prompt for automated scoring is available in Appendix~\ref{appendix_punc_itn}. The first place receives 3 points, the second place 2 points, and the third place 1 point. The final score is the average of all scores. When preparing the test data, we use randomized option shuffling and multiple scoring rounds to reduce uncertainty when using ChatGPT for evaluation. The final results demonstrate that \textbf{MinMo performs better in the subjective evaluations of punctuation insertion and ITN}.

\subsection{Voice Generation}
\label{sec:voice_gen}
\paragraph{Text-to-Speech (TTS)} To evaluate the synthesis accuracy of our voice decoder, we converted the recent SEED test set \citep{seed-tts} into the ChatLM format. In this format, the text is presented as the user content prefixed with a ``Copy:'' command, and the LLM is expected to replicate this text.
The test set comprises 2,020 cases in Chinese and 1,088 cases in English.
For the Chinese cases, we utilized the Paraformer-zh model \citep{Paraformer}, while the English cases were processed using Whisper-large V3 \citep{radford2023robust}.
Given the instruction non-following issue with LLMs, we applied a teacher forcing scheme during inference to minimize discrepancies between the input and output text. The content consistency of the voice decoder was evaluated using CER for Chinese and WER for English.

Our findings indicate that even with the teacher forcing scheme, only about 20\% of the test cases had identical input and output text from the LLM. Because inconsistent input and output can lead to confused hidden states for the voice decoder, only test cases with consistent input-output text were included for error rate calculation.
The results are presented in Table \ref{tab:tts}. 
We observed that MinMo's voice decoder has slightly reduced content consistency and speech quality on the Chinese test sets compared to the TTS baseline, CosyVoice 2.0-SFT \citep{du2024cosyvoice2scalablestreaming}. On the English test set, MinMo achieves similar content consistency but with a slightly lower NMOS score. This reduction can be attributed to the differing acoustic characteristics of the fine-tuned speakers, which affect both the recognition model and NMOS scorer. However, this reduction does not significantly hinder human understanding. Therefore, subjective evaluation might be more appropriate for speech-to-speech voice chat models, which will be explored in our future work.

\begin{table}[htbp]
\centering
\begin{tabular}{l cc cc}
\toprule
\multirow{2}{*}{\textbf{Model}} & \multicolumn{2}{c }{\textbf{zh}} & \multicolumn{2}{c}{\textbf{en}} \\
\cmidrule{2-5}
& \textbf{CER} & \textbf{NMOS} & \textbf{WER} & \textbf{NMOS} \\
\midrule
CosyVoice 2.0-SFT & 2.06 & 3.73 & 3.19 & 3.71 \\
MinMo & 2.48 & 3.69 & 2.90 & 3.56 \\
\bottomrule
\end{tabular}
\vspace{3mm}
\caption{Performance comparison of content consistency (CER/WER) and objective speech quality (NMOS) between MinMo and the TTS baseline CosyVoice 2.0-SFT on the text-to-speech Chinese (zh) and English (en) test sets.}
\label{tab:tts}
\end{table}

\paragraph{Instruction-following Voice Generation}

To evaluate the performance of instruction-following voice generation, we develop a multi-turn Chinese speech-to-speech test set consisting of 30 sessions and 122 turns, incorporating 12 types of instructional controls. These controls include emotions (happy, sad, surprised, angry, fearful), dialects (Cantonese, Sichuan), speaking rates (fast, slow), role-playing (robot, Peppa), and a default style.
To assess the accuracy of instruction-following voice generation, listeners classify the generated audio according to the instruction type. As shown in Table \ref{tab:res-instruct}, MinMo demonstrates superior instruction control accuracy compared to the baseline GLM-4-Voice, particularly in dialects and role-playing.

\begin{table}[htbp]
\centering
\begin{tabular}{ccccccc}
\toprule
Model & Emotion & Dialect & Speaking Rate & Role-playing & Default & Total \\ 
\midrule
GLM-4-Voice & 75.6  & 42.9 & 80.0 & 70.4 & 88.2 & 63.1 \\
MinMo & 97.6 & 100 & 100 & 96.3 & 96.3 & 98.4  \\ 
\bottomrule
\end{tabular}
\vspace{3mm}
\caption{Performance comparison of instruction-following voice generation between MinMo and the baseline GLM-4-Voice on the multi-turn speech-to-speech Chinese test set.}
\label{tab:res-instruct}
\end{table}

\subsection{Voice Chat}
\paragraph{Spoken Question Answering and Spoken Dialogue}
To transfer the dialog capabilities of the base model to the speech modality, we construct multi-turn conversational data for both speech-to-text (speech2text) and speech-to-speech (speech2speech) scenarios. The speech2text data is primarily divided into two parts. First, it originates from open-source multi-turn text-only data, where we synthesize the user turns using zero-shot Text-to-Speech (TTS) technology. Second, we use real Automatic Speech Recognition (ASR) training data as chat queries to obtain text responses from the large model, thereby generating interactive training data for speech2text.

To evaluate the question-answering capabilities of MinMo in the speech modality, we initially utilize three datasets, namely Llama Questions ~\citep{nachmani2024spokenquestionansweringspeech}, Trivia QA ~\citep{joshi2017triviaqalargescaledistantly}, and Web Questions ~\citep{DBLP:conf/emnlp/BerantCFL13}, similar to the approach by \citet{DBLP:journals/corr/abs-2410-00037}, \citet{zeng2024glm} and \citet{wang2024freeze}. These datasets are employed to assess the model's knowledge question-answering ability in both speech-to-text and speech-to-speech modes. The baseline results for these datasets are taken from the original table of the respective papers. It is noteworthy that, due to the absence of a publicly available specific test set for the Trivia QA, we adhere to the GLM configuration by randomly selecting 1,000 samples from the validation set of web questions as our test set. However, because of the inconsistency in test samples, the results on Trivia QA are not guaranteed to be absolutely comparable and should be considered as reference only. Additionally, since the original format of these three datasets is text-based, during testing, we use CosyVoice2\citep{du2024cosyvoice2scalablestreaming} to perform TTS synthesis on the input question texts.

As illustrated in the table~\ref{tab:s2t_chat_moshi}, the MinMo model demonstrates a significant advantage over existing baselines in the Speech-to-Speech (S2S) mode, achieving new state-of-the-art (SOTA) results. In the Speech-to-Text (S2T) mode, it also attains SOTA performance on the Llama Question and Web Question datasets. However, the test results of MinMo still indicate a noticeable performance decline in the S2S mode compared to the S2T mode. We attribute this to the fact that many answers in the test set are rich in textual structure and specialized vocabulary, which imposes greater demands on the model's text-to-speech (TTS) capabilities. Additionally, the automatic speech recognition (ASR) model used to obtain the answers text for speech in the S2S evaluation can also impact the S2S metrics to some extent.

In order to further analyze MinMo's voice interaction capabilities, we additionally constructed two test sets. The evaluation criteria are divided into two parts: first, assessing the preservation of MinMo's logical reasoning abilities in the speech modality; second, evaluating MinMo's spoken response capabilities in casual voice interactions. To facilitate this analysis, we construct two evaluation test sets: the Alpaca test set~\citep{alpaca_eval}, which emphasizes logical reasoning capabilities, and the ChitChat test set, which targets casual conversational scenarios. Referencing the work of~\citet{DBLP:conf/nips/ZhengC00WZL0LXZ23}, the evaluation capabilities of current high-quality large models align well with human assessments. Therefore, to enhance evaluation efficiency, we employ automated scoring using large models. The specific task prompt for the automated scoring can be found in the appendix~\ref{sec:appendix}. We used Qwen-Max as the scoring model, with each dialogue sample receiving a score ranging from 0 to 10. The average score of the samples is taken as the final score.

From Table~\ref{tab:s2t_chat_exp}, it can be observed that by incorporating additional speech2text task data into MinMo training, we are able to effectively maintain the conversational capabilities of the base model. Compared to the performance of the ASR combined with the text-only base model, MinMo's conversational ability remains largely consistent. However, MinMo's response scores are slightly lower than the quality of Ground Truth responses. We believe this discrepancy can be attributed to two main reasons. Firstly, the integration of multiple speech tasks and the implementation of LoRA training on the base model have somewhat diminished the logical generation capabilities of the original Large Language Model (LLM). The table shows that, compared to the ChitChat test set, MinMo exhibits greater performance variations on the Alpaca test set. Second, there is room for further improvement in MinMo's audio comprehension capabilities, and there remains potential for reducing the Character Error Rate (CER) in ASR tasks.

\begin{table}[htbp]
\centering
\begin{tabular}{ccccccc}
\toprule
\multirow{2}{*}{Model} & \multicolumn{2}{c}{Llama Question} & \multicolumn{2}{c}{TriviaQA{$^*$}} & \multicolumn{2}{c}{Web Questions} \\
 & S2T & S2S & S2T & S2S & S2T & S2S \\ \midrule
Moshi~\citep{DBLP:journals/corr/abs-2410-00037} & 62.3 & 21 & 22.8 & 7.3 & 26.6 & 9.2 \\
GLM-4-Voice~\citep{zeng2024glm} & 64.7 & 50.7 & 39.1 & 26.5 & 32.2 & 15.9 \\
Freeze-Omni~\citep{wang2024freeze} & 72.0 & - & 53.9 & - & 44.7 & - \\
MinMo & \textbf{78.9} & \textbf{64.1} & 48.3 & 37.5 & \textbf{55.0} & \textbf{39.9} \\ \bottomrule
\end{tabular}
\vspace{3mm}
\caption{Comparison of Spoken Question Answering Performance: Results for Moshi, GLM-4-Voice, and Freeze-Omni are sourced from their respective papers. S2T refers to the Speech-to-Text evaluation, while S2S denotes the Speech-to-Speech evaluation. The metric used for these assessments is accuracy. \textit{The TriviaQA$^*$ dataset does not provide a public test set, so the numerical results are not directly comparable and should be considered for reference only.}}
\label{tab:s2t_chat_moshi}
\end{table}

\begin{table}[htbp]
\centering
\begin{tabular}{ccc}
\toprule
 & Alpaca Test~\citep{alpaca_eval} & ChitChat Test \\ 
\midrule
Ground Truth & 7.73 & 7.62 \\
ASR + Qwen2.5 & 6.59 & 7.18 \\
MinMo & 6.48 & 7.20 \\ 
\bottomrule
\end{tabular}
\vspace{3mm}
\caption{Performance of MinMo in two in-house multi-turn speech-to-speech test sets: the Alpaca test set and the ChitChat test set. The Alpaca test set focuses on assessing logical reasoning capabilities, while the ChitChat test set is designed to evaluate casual conversational scenarios. For scoring, we utilized the Qwen-Max model.}
\label{tab:s2t_chat_exp}
\end{table}
\paragraph{Full Duplex Spoken Dialogue}
To assess the capabilities of MinMo in full-duplex voice interaction, we construct three test sets: the Chinese Alimeeting dataset, the English Fisher dataset, and a simulated test set designed to more closely resemble real human-machine dialogue scenarios. We evaluate MinMo's full-duplex capabilities from two perspectives: prediction performance and prediction efficiency. Regarding prediction performance, the evaluation is divided into three tasks: assistant turn-taking, user turn-taking, and user back-channeling. For the turn-taking tasks, we employe the positive F1 score as our analytical metric and also introduced the offset distance ( K ) to better analyze the model's performance. For the user back-channel task, we utilize accuracy to assess MinMo's ability to recognize back-channel utterances.

From Table \ref{tab:fd_performance}, it can be observed that the MinMo model demonstrates commendable results on the human-machine conversation dataset, irrespective of whether it is user turn-taking or assistant turn-taking. At K=10, the prediction performance approaches 99\%. On the test set of actual human-human conversations, the performance of the MinMo model on assistant turn-taking shows a certain degree of decline compared to the human-machine conversation test set. We believe this is primarily due to the high variability in background noise, speech speed, pauses, and other factors in real human conversations, which can lead to some degree of misjudgment by the model on the assistant turn-taking task. However, for user turn-taking prediction in human-human conversations, the MinMo model still maintains a high level of sensitivity and predictive performance, ensuring the system promptly stops speaking when the user talks, thereby avoiding overlapping speech with the user. This sensitivity and respect for user speech also explain why the MinMo model maintains a prediction accuracy of 70\%-80\% for user back-channel comments, as shown in the table. This is consistent with the tuning of the user turn-taking model, indicating a certain trade-off between the two.

For the efficiency analysis of the MinMo duplex mode, we also conduct tests separately on both human-human dialogue and human-machine dialogue test sets. As shown in Table \ref{tab:fd_efficiency}, the average response delay of MinMo in user turn-taking is 250ms. The fastest response speed is observed in the human-machine test set, at 88.8ms, while the most challenging Alimeeting test set shows a delay of 448.8ms. In terms of assistant turn-taking, the average response delay of MinMo is around 660ms, which is longer compared to the response time required for user turn-taking prediction. We attribute this to the fact that user turn-taking involves the beginning part of the user's speech, whereas assistant turn-taking involves the part where the user's turn is nearly finished. Therefore, the contextual semantic information for assistant turn-taking is more comprehensive, which results in a shorter time lag needed for decision-making.

\paragraph{Full Duplex System Latency}
The duplex interaction of MinMo consists of four modules: the Full-duplex predictor, responsible for duplex control, the speech-to-text module (Voice Encoder+Input Projector+LLM), the text-to-speech token module (Output Projector+Voice Token LM), and the Token2Wav module. The latencies for each module are shown in Table \ref{table:latency_analysis}. Taking Assistant Turn-taking as an example, when the user's actual speech concludes, the duplex model typically requires a delay of 250 ms for evaluation. The prediction of the initial five text tokens in the Speech-to-Text process takes approximately 150 ms. Predicting the initial 15 speech tokens requires about 70 ms, and transitioning from speech tokens to the first audio packet takes an additional 130 ms. Consequently, when developing a full-duplex voice dialogue system based on MinMo, the standard experiential delay for assistant turn-taking is approximately 250 + 150 + 70 + 130 = 600 ms. The aforementioned numerical estimates are derived during testing using the L20 GPU and BF16 model format.
\begin{table}[htbp]
\centering
\begin{tabular}{ c c c c }
\toprule
\textbf{Full-duplex predictor} & \textbf{Speech-to-text} & \textbf{Text-to-speech token} & \textbf{Token2Wav} \\
       (Assistant turn-taking)  &(1 or 5 text tokens)       & (15 speech tokens)            &                     
\\ 
\midrule
250ms & 95ms or 150ms & 70ms & 130ms \\ 
\bottomrule
\end{tabular}
\vspace{3mm}
\caption{MinMo's system latency analysis in two L20 GPUs.}
\label{table:latency_analysis}
\end{table}

\begin{table}[htbp]
\centering
\begin{tabular}{lccc}
\toprule
\multicolumn{1}{c}{Data} & \begin{tabular}[c]{@{}c@{}}Assistant’s Turn-taking \\ (Pos. F1, @K=1/5/10)\end{tabular} & \begin{tabular}[c]{@{}c@{}}User's Turn-taking \\ (Pos. F1, @K=1/5/10)\end{tabular} & \begin{tabular}[c]{@{}c@{}}User's Back-channel \\ (Acc. )\end{tabular} \\ \midrule
Alimeeting & 0.6138 / 0.7542 / 0.8036 & 0.4751 / 0.9366 / 1 & 0.7124 \\
Fisher & 0.6682  / 0.8372 / 0.8813 & 0.4271 / 0.9455 / 0.9994 & 0.8123 \\
Simulation & 0.7868 / 0.9616 / 0.985 & 0.2571 / 0.8152 / 0.9942 & - \\ \bottomrule
\end{tabular}
\vspace{3mm}
\caption{Performance evaluation of the duplex prediction module: the turn-taking between assistant and user is measured using the positive F1 score @offset-K metric, while the user's back-channel is assessed using the accuracy metric.}
\label{tab:fd_performance}
\end{table}

\begin{table}[htbp]
\centering
\begin{tabular}{lcc}
\toprule
\multicolumn{1}{c}{\multirow{2}{*}{Data}} & \multicolumn{2}{c}{Average Latency  (ms.)} \\ \cmidrule(lr){2-3} 
\multicolumn{1}{c}{} & \multicolumn{1}{l}{Assistant's Turn-taking} & \multicolumn{1}{l}{User's Turn-taking} \\ \midrule
Alimeeting & 448.8 & 663.4 \\
Fisher & 189.1 & 641.8 \\
Simulation & 88.8 & 673.7 \\ \bottomrule
\end{tabular}
\vspace{3mm}
\caption{Average latency (ms) of MinMo for assistant's turn-taking and user's turn-taking on various full-duplex voice chat data sets.}
\label{tab:fd_efficiency}
\end{table}

\section{Conclusion}
\label{sec:conclusion}
This research introduces MinMo, an advanced multimodal large language model designed to overcome the limitations of existing aligned multimodal models in seamless voice interaction. Trained on an extensive dataset of over 1.4 million hours of speech, MinMo showcases state-of-the-art performance across diverse benchmarks, including spoken dialogue, multilingual speech recognition, and emotion recognition. By leveraging a multi-stage alignment strategy, MinMo adeptly balances audio understanding and generation while minimizing the catastrophic forgetting often observed in text-based LLMs.
A key innovation is MinMo's novel alignment method for streaming end-to-end audio generation. By utilizing hidden layer representations of the text model, MinMo’s voice decoder achieves structural simplicity and competitive performance with low latency. This approach significantly enhances the model’s instruction-following capabilities, enabling nuanced speech generation that accurately reflects user-specified emotions, dialects, and speaking styles. Furthermore, MinMo supports full-duplex interactions, facilitating a seamless conversational experience with a latency of approximately 600ms.
In conclusion, MinMo represents a substantial advancement in the field of voice interaction systems. It not only addresses the inherent challenges of sequence length discrepancies and data imbalance but also sets a new standard for natural and expressive voice interactions, paving the way for future developments in multimodal language models.

\section{Limitations}
MinMo has certain limitations that need to be addressed. Firstly, MinMo integrates audio understanding and audio generation capabilities based on a pre-trained text large model by using alignment. The text large model only participates in LoRA updates, and its ability to follow diverse instructions, such as language and task following, needs improvement. Further exploration is needed to determine whether using more high-quality text data for more comprehensive updates of the text large model can enhance its instruction-following ability. Secondly, there are some long-tail pronunciation error issues in MinMo's end-to-end audio generation. This problem partly arises from retaining some one-to-many tokens of the LLM, and partly because some special symbols in the end-to-end modeled output text cannot be effectively converted into speech. Data scaling can be explored to address these long-tail issues. Additionally, the overall efficiency of audio generation controlled by instructions in MinMo needs to be improved. This is partly due to the overall small size of the current instruction data and the limitation of only using hidden embeddings for end-to-end alignment, which restricts the transmission of historical information. Finally, while MinMo implements a duplex module based on semantics, it still requires separate AEC and VAD modules. In the future, a fully end-to-end duplex model will be explored.

\section{Authors (alphabetical order of family name)}
\begin{multicols}{3}
  \begin{itemize}[noitemsep]
    \item Qian Chen
            \item Yafeng Chen
            \item Yanni  Chen
            \item Mengzhe Chen
            \item Yingda Chen
            \item Chong Deng
    \item Zhihao Du
            \item Ruize Gao
    \item Changfeng Gao
    \item Zhifu Gao
            \item Yabin Li
    \item Xiang Lv
            \item Jiaqing Liu
            \item Haoneng Luo
            \item Bin Ma
            \item Chongjia Ni
            \item Xian Shi
            \item Jialong Tang
            \item Hui Wang
            \item Hao Wang
    \item Wen Wang
    \item Yuxuan Wang
            \item Yunlan Xu
            \item Fan Yu
    \item Zhijie Yan
    \item Yexin Yang
            \item Baosong Yang
            \item Xian Yang
            \item Guanrou Yang
            \item Tianyu Zhao
    \item Qinglin Zhang
    \item Shiliang Zhang
    \item Nan Zhao
            \item Pei Zhang
            \item Chong Zhang
            \item Jinren Zhou
  \end{itemize}
\end{multicols}

\section{Acknowledgment}

We are immensely grateful for the invaluable discussions, support, and assistance we received from many colleagues during the development and demonstration of the MinMo model. Special thanks go to:
Keyu An,
Cheng Chen,
Luyao Cheng,
Rui Li,
Jiayi Li,
Minjun Liang,
Chaohong Tan,
Yiwei Wang,
Ming Zhao.

Additionally, we appreciate to the ModelScope community for providing GPU computational support for the demo services. Their contributions and support have been crucial in bringing this project to fruition.

\bibliographystyle{iclr2023_conference}
\bibliography{refs}

\appendix
\label{sec:appendix}
\section{Prompts for Voice Understanding Tasks}
\label{appendix_voice_understanding_task_prompts}
\subsection{Spoken Language Smoothing}
\label{appendix_spoken_Language_smoothing}

Here is the prompt for faithfulness and formality evaluation of the spoken language smoothing task, taking meetings as an example.

\begin{tcolorbox}[colback=green!5!white, colframe=green!80!black, title=Prompt for Faithfulness Evaluation (Spoken Language Smoothing)]
Suppose you are a professional text editor, please evaluate the quality of the model's refined output. The original content is a paragraph from Automatic Speech Recognition (ASR), which may contain recognition errors, grammatical errors, and spoken/informal style expressions. The model's refined output should be free of noticeable ASR errors and grammatical errors, faithfully preserve the original content, and exhibit a clear written style with excellent readability. The human's refined output can be considered as one of the qualified polished result, which can be used as a reference for quality assessment. Please evaluate and score the model's refined output. Use a scoring range from 1 to 10 for each criterion, where 1 represents the worst performance and 10 represents the best performance. The return fields include:

1. ASR Error Correction Score: Check whether the text effectively corrects errors from automatic speech recognition. The higher the score, the more accurate the correction.

2. Faithful Score: Determine whether the text polishing results maintain the intent and content of the original speech. The higher the score, the more complete the preservation of the original meaning.

3. Evaluation Summary: Based on the scores above, briefly explain the reasons for the scores, provide an overall evaluation of the polishing quality, and suggest modifications.

The following is the original content:

\{source\}

The following is the human's refined output for reference:

\{human-target\}

The following is the model's refined output to evaluate:

\{model-target\}

Please provide scores for each criterion and an overall evaluation with json format:

\end{tcolorbox}

\begin{tcolorbox}[colback=green!5!white, colframe=green!80!black, title=Prompt for Formality Evaluation (Spoken Language Smoothing)]
Suppose you are a professional text editor, please evaluate the quality of the model's refined output. The model's refined output should be free of grammatical errors, and exhibit a clear written style with excellent readability. Please evaluate and score the model's refined output. Use a scoring range from 1 to 10 for each criterion, where 1 represents the worst performance and 10 represents the best performance. The return fields include:

1. Grammar Score: Evaluate the grammatical accuracy of the text. The higher the score, the fewer grammatical errors there are.

2. Formal Score: Assess the written language style of the text, including the level of formality and suitability for written communication. The higher the score, the stronger the written style of the text.

3. Readability Score: Evaluate the fluency and overall readability of the text, including ease of understanding and natural expression. The higher the score, the better the readability of the text.

4. Evaluation Summary: Based on the scores above, briefly explain the reasons for the scores, provide an overall evaluation of the polishing quality, and suggest modifications.

The following is the model's refined output to evaluate:

\{target\}

Please provide scores for each criterion and an overall evaluation with json format:

\end{tcolorbox}

\subsection{Punctuation and Inverse Text Normalization}
\label{appendix_punc_itn}

Here is the prompt for evaluation of punctuation and inverse text normalization.

\begin{tcolorbox}[colback=green!5!white, colframe=green!80!black, title=Prompt for Faithfulness Evaluation (Spoken Language Smoothing)]
Compare the punctuation usage in the following sentences (focus only on punctuation, ignore other differences and the issue of full-width and half-width characters). Make a ranking and output only the results of the ranking without any explanation. Display the results in the order of ranking, for example, a:1; b:3; c:2, where 1 represents first place, 2 represents second place, and so on. Ranks can be the same. The options are as follows:

\{option1\}

\{option2\}

\{option3\}
\end{tcolorbox}

\begin{tcolorbox}[colback=green!5!white, colframe=green!80!black, title=Prompt for Formality Evaluation (Spoken Language Smoothing)]
Compare the inverse text normalization in the sentences below, focusing only on the aspect of inverse text normalization and not on differences in the text or punctuation itself. Make a ranking and output only the results of the ranking without any explanation. Display the results in the order of ranking, for example, a:1; b:3; c:2, where 1 represents first place, 2 represents second place, and so on. Ranks can be the same. The options are as follows:

\{option1\}

\{option2\}

\{option3\}
\end{tcolorbox}

\end{document}